%% file: main.tex
\documentclass[nohyperref]{article}

\usepackage{microtype}
\usepackage{graphicx}
\usepackage{booktabs} 

\usepackage{hyperref}

\usepackage[accepted]{icml2022}

\usepackage{amsmath}
\usepackage{amssymb}
\usepackage{mathtools}
\usepackage{amsthm}
\usepackage{url}
\usepackage{amsfonts}       
\usepackage{nicefrac}       
\usepackage{xcolor}         
\usepackage{times}
\usepackage{epsfig}
\usepackage{pifont}
\usepackage{kotex}
\usepackage{enumitem}
\usepackage{verbatim}
\usepackage{bm}
\usepackage{makecell,multirow, tabularx}
\usepackage{algorithm}
\usepackage{algorithmic}
\usepackage{url}
\usepackage{balance}
\usepackage{upquote}
\usepackage{appendix}
\usepackage{nicefrac}
\usepackage{physics}
\usepackage{stmaryrd}
\usepackage{pifont}
\usepackage{wrapfig} 
\DeclareMathOperator*{\argmax}{arg\,max}
\DeclareMathOperator*{\argmin}{arg\,min}
\usepackage{caption}
\usepackage{subcaption}

\usepackage{xcolor}
\usepackage{xspace}
\newcommand\eg{\textit{e.g.}\xspace}
\newcommand\ie{\textit{i.e.}\xspace}

\usepackage[capitalize,noabbrev]{cleveref}

\theoremstyle{plain}

\theoremstyle{definition}

\theoremstyle{remark}

\usepackage[textsize=tiny]{todonotes}

\icmltitlerunning{Dataset Condensation With Contrastive Signals}

\begin{document}

\twocolumn[
\icmltitle{Dataset Condensation with Contrastive Signals}



\icmlsetsymbol{wdan}{\dag}
\begin{icmlauthorlist}
\icmlauthor{Saehyung Lee}{snu,wdan}
\icmlauthor{Sanghyuk Chun}{naver}
\icmlauthor{Sangwon Jung}{snu,wdan}
\icmlauthor{Sangdoo Yun}{naver}
\icmlauthor{Sungroh Yoon}{snu,ipai}
\end{icmlauthorlist}

\icmlaffiliation{snu}{Department of Electric and Computer Engineering, Seoul National University}
\icmlaffiliation{naver}{NAVER AI Lab}
\icmlaffiliation{ipai}{Interdisciplinary Program
in Artificial Intelligence, Seoul National University}

\icmlcorrespondingauthor{Sungroh Yoon}{\href{mailto:sryoon@snu.ac.kr}{sryoon@snu.ac.kr}}

\icmlkeywords{Machine Learning, ICML}

\vskip 0.3in
]



\printAffiliationsAndNotice{\icmlIntern} 

\input{abstract}
\input{introduction}
\input{related_work}
\input{method}
\input{experiments}
\input{conclusion}

\paragraph{Acknowledgements:}
Most experiments were conducted on NAVER Smart Machine Learning (NSML) platform \cite{nsml1, nsml2}.
This work was supported by the National Research Foundation of Korea (NRF) grant funded by the Korea government (MSIT) (No. 2022R1A3B1077720),
Institute of Information \& communications Technology Planning \& Evaluation (IITP) grant funded by the Korea government(MSIT) [NO.2021-0-01343, Artificial Intelligence Graduate School Program (Seoul National University)], and the BK21 FOUR program of the Education and Research Program for Future ICT Pioneers, Seoul National University in 2022.

\balance
\bibliography{reference}
\bibliographystyle{icml2022}

\newpage
\nobalance
\appendix
\onecolumn
\input{appendix}

\end{document}

%% file: abstract.tex
\begin{abstract}
Recent studies have demonstrated that gradient matching-based dataset synthesis, or dataset condensation (DC), methods can achieve state-of-the-art performance when applied to data-efficient learning tasks. However, in this study, we prove that the existing DC methods can perform worse than the random selection method when task-irrelevant information forms a significant part of the training dataset. We attribute this to the lack of participation of the contrastive signals between the classes resulting from the class-wise gradient matching strategy.
To address this problem, we propose Dataset Condensation with Contrastive signals (DCC) by modifying the loss function to enable the DC methods to effectively capture the differences between classes. In addition, we analyze the new loss function in terms of training dynamics by tracking the kernel velocity. Furthermore, we introduce a bi-level warm-up strategy to stabilize the optimization. Our experimental results indicate that while the existing methods are ineffective for fine-grained image classification tasks, the proposed method can successfully generate informative synthetic datasets for the same tasks. Moreover, we demonstrate that the proposed method outperforms the baselines even on benchmark datasets such as SVHN, CIFAR-10, and CIFAR-100. Finally, we demonstrate the high applicability of the proposed method by applying it to continual learning tasks.
\end{abstract}

%% file: introduction.tex
\section{Introduction}\label{introduction}
Deep neural networks (DNNs) are data hungry; larger datasets make DNNs more generalizable (\eg, by data augmentation \cite{mixup, cutmix, oat}, or by collecting hyperscale training datasets \cite{align_google}). Unsurprisingly, gigantic datasets (\eg, 410 B language tokens \cite{gpt3}, 3.5B images \cite{instagramnet}, and 1.8B image-text pairs \cite{align_google}) have become central to the training of ground-breaking deep models. However, such large datasets require tremendous computational and infrastructural resources, not only for training deep models but also for collecting and processing data columns. Furthermore, real-world knowledge is increasing exponentially, while machine learning (ML) systems are prone to catastrophic forgetting \cite{catastrophic_forgetting,rebuffi2017icarl}. This necessitates repeated training using massive training samples to ensure that ML applications remain competent and practical. Thus, considering the high computational costs, dataset reduction methods are extremely beneficial in applied ML fields.
\begin{figure}[t]
    \centering
    \includegraphics[width=\linewidth]{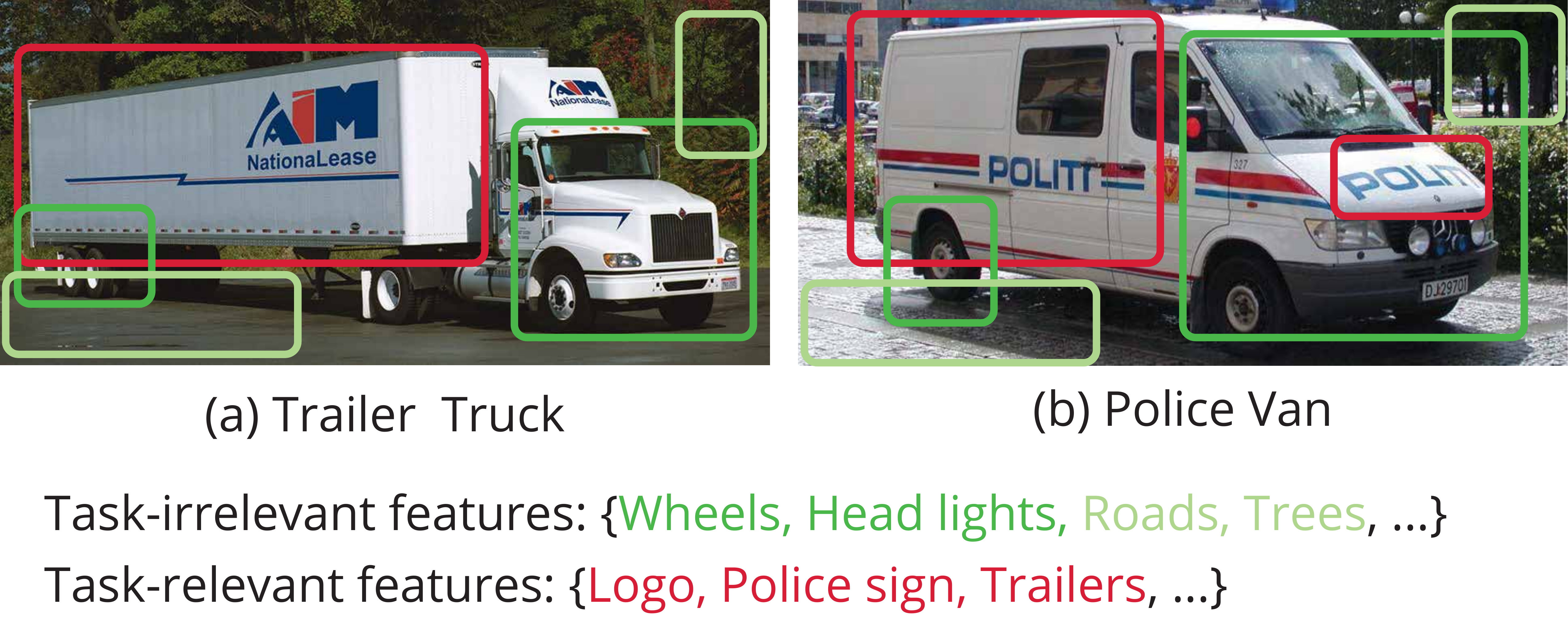}
    \caption{Example of a fine-grained Truck classification task where the task-irrelevant common features are dominant and the task-relevant discriminant features are in the minority.}
    \label{fig:finegrained_contrastive_signal}
\end{figure}
\citet{dc} proposed a \emph{dataset condensation} method (DC) to synthesize a small but informative dataset by matching the loss gradients with respect to the training and synthetic datasets.
In particular, DC was developed to be suitable for downstream classification tasks by repeating the classifier training during the synthetic data optimization procedure,
resulting in reasonable performances with reduced synthetic datasets.

In this study, however, we show that DC primarily focuses on the class-wise gradient while overlooking \emph{contrastive signals}. Thus, DC underperforms even when compared with the random selection baseline when contrastive signals are significant to the task. For example, in fine-grained image classification tasks, such as Truck categorization in Fig. \ref{fig:finegrained_contrastive_signal}, contrastive signals should be considered to encode task-relevant information (\eg, logo, police sign, trailers) while suppressing task-irrelevant information (\eg, wheels, head lights, roads, trees).
In our experiments on the fine-grained Automobile dataset, DC results in a classifier with a test accuracy (11\%) lower than that achieved using the random selection method (12.2\%).
We demonstrate that DC cannot effectively utilize the contrastive signals of interclass samples using a motivating example and qualitative analysis.

To address this issue, we propose the \emph{Dataset Condensation with Contrastive signals (DCC)} method. This introduces a modified gradient matching loss function that enables the optimization of a synthetic dataset to capture the contrastive signals. In contrast to DC, which employs only training data of the same class when synthesizing images for a specific class by the class-wise gradient matching, DCC matches the sum of gradients over all classes with respect to the synthetic and training datasets. Additionally, we analyze our method in terms of training dynamics by tracking the kernel velocity \cite{kernel_velocity} and introduce a bi-level warm-up strategy to stabilize the optimization procedure of our method.
In our experiments, we demonstrate that the proposed DCC singularly outperforms DC in fine-grained classification tasks and general benchmark datasets, such as SVHN \cite{svhn_dataset}, CIFAR-10, and CIFAR-100 \cite{cifar_dataset}. Finally, we also demonstrate the superiority of our method compared with baselines on downstream tasks, where small synthetic datasets efficiently reduce the total storage of data (\eg, continual learning).
The code of our study is available at: \url{https://github.com/Saehyung-Lee/DCC}

%% file: related_work.tex
\section{Related Work}\label{related work}
Formally, we define \emph{the dataset reduction problem} as follows:
\begin{equation}\label{eq:data_reducing_problem}
    \mathcal{S}^\star=\argmax_{\mathcal{S}} I(\mathcal{X};\mathcal{S}\mid\tau).
\end{equation}
Here, $\mathcal{X}=\{X_n\}_{n=1}^N$ and $\mathcal{S}=\{S_k\}_{k=1}^K$ are the training and reduced datasets, respectively, and $K \ll N$. $\tau$ is a task-dependent variable, and $I(\mathcal{X};\mathcal{S}\mid\tau)$ is the conditional mutual information.
Herein, we focus on classification tasks, which is the most widely studied scenario with respect to dataset reduction tasks \cite{dd,dc}.

\paragraph{Selection-based methods.}
Selection-based methods \cite{coresets_for_efficient} find a data subset (coreset) that satisfies the cardinality constraint (\ie, $|\mathcal{S}|=K$) while minimizing the difference between the loss gradient on the training dataset and that on the coreset. Moreover, recent studies \cite{cscores,diet} have demonstrated that a large fraction of training dataset can be pruned based on the scores they provide.
\citet{cscores} introduced a consistency score (C-score) that represents the expected accuracy for a held-out sample on a training dataset. By sorting the samples according to their C-scores, we can identify prototypical (high-scoring) samples that can serve as a proxy for the training dataset.
\citet{diet} proposed
the use of the expected loss gradient norm (GradNd) or the norm of the error vector (EL2N) of each training sample to prune a fraction of the training samples. In our case, we preserve the most typical (low-scoring) samples to obtain a proxy for the training dataset.

The selection-based methods, however, are ineffective, particularly when the task-conditional data information $H(\mathcal{X}\mid\tau)$ is evenly divided and distributed among the training samples. To be precise, if $H(X_n\mid\mathcal{X} \setminus \{X_n\}, \tau)=\frac{1}{N}H(\mathcal{X}\mid\tau)$ for all $n\in\{1, \dots, N\}$, then the mutual information $I(\mathcal{X};\mathcal{S}^\star\mid\tau)$ found by any selection-based method is always a small value $\frac{K}{N}H(\mathcal{X}\mid\tau)$. As shown by \citet{dc}, the empirical performance gaps between the existing data selection methods and random selection baselines are of no significance in most realistic evaluation benchmarks. 

\paragraph{Synthesis-based methods.}
Instead of selecting a subset from the training dataset, a small dataset $\mathcal{S}$ that achieves similar performance to $\mathcal{X}$ can be generated. Ideally, assuming that the capacity of a synthetic datum can contain as much information as $\frac{1}{K}H(\mathcal{X}\mid\tau)$, there exists an $\mathcal{S}$ that achieves the same task performance as that attained through the use of $\mathcal{X}$.
\citet{dd} proposed dataset distillation (DD) to transfer the knowledge from a large dataset to a small dataset. They demonstrated that it is possible to achieve close to original accuracy on MNIST \cite{mnist_dataset} using merely ten synthetic images.
Inspired by DD, \citet{dc} proposed a DC to synthesize a small set of informative samples for learning downstream tasks. The authors showed that the DC outperformed all the baselines in their experiments.
Recently, \citet{kip} proposed a meta-learning algorithm called Kernel Inducing Points (KIP) for the dataset reduction problem. Furthermore, they presented state-of-the-art performance by using infinitely wide convolutional neural networks \citep{dd_infinite}.

%% file: Method.tex
\section{Method}\label{method}
In this section, we introduce the DC method \cite{dc} (Sec. \ref{subsec:dc}), and study a motivating example showing the limitation of the class-wise gradient matching strategy employed by DC (Sec. \ref{subsec:motivating_example}). To mitigate this issue, we propose a modified gradient matching loss method (Sec. \ref{subsec:dcc}).
Furthermore, we propose a bi-level warm-up strategy to stabilize the optimization of the proposed loss function.


\subsection{Preliminary: DC with Gradient Matching}\label{subsec:dc}
When generating images for class ``$c$'' ($\mathcal{S}^c$), DC uses only the training data of class ``$c$'' ($\mathcal{X}^c$). In particular, DC first (\romannumeral 1) updates a synthetic dataset $\mathcal{S}$ by applying a gradient descent step toward the minimization of the following loss $\mathcal{L}$:
\begin{equation} \label{eq:dc}
    \mathcal{L}=\sum_{c=0}^{C-1}D\left( \nabla_{\theta_t}L(\mathcal{X}^c;\theta_t), \nabla_{\theta_t}L(\mathcal{S}^c;\theta_t) \right),
\end{equation}
where $C$, $D(\cdot,\cdot)$ and  $L(\cdot;)$ denote the number of classes, a distance function, and the cross-entropy loss function, respectively, and $\nabla_{\theta_t}L(\mathcal{X}^c)$ is the average loss gradient with respect to a model $\theta_t$; (\romannumeral 2) before moving on to step $t+1$, trains the model on $\mathcal{S}$; (\romannumeral 3) alternately optimizes the synthetic dataset and the model; and (\romannumeral 4) randomly initializes the model after every pre-defined period $T$ (\ie, $\{\theta_{iT}\mid i\in\mathbb{N}_0\}$ is a set of randomly initialized models). Periodic model initialization plays an important role in ensuring that $\mathcal{S}$ can be used for previously unseen models. In addition, \citet{dsa} improved DC by using Differentiable Siamese Augmentation (DSA) to generate more informative synthetic datasets. DSA transforms both $\mathcal{X}^c$ and $\mathcal{S}^c$ with the same random transformation (\eg, color jittering, cropping, cutout, flipping, and scale) at each training step. Except for the transformation part, the DSA and DC methods are identical. That is, DSA also uses class-wise gradient matching loss and periodic model initialization.

\subsection{A Motivating Example}\label{subsec:motivating_example}
In this subsection, we show an example, in which the class-wise gradient matching strategy (employed by DC) is problematic. In particular, we show that the class-wise gradient matching strategy is dominated by task-irrelevant \emph{class-common} features, whereas the \emph{class-discriminative} features are relatively neglected. Fig. \ref{fig:motivating_example} presents an overview.

\paragraph{Setup.}
\input{Figure/figure_motivating_example}
We define a binary classification dataset $\mathcal{X}=\{(x_n,y_n)\}_{n=1}^{N}$ sampled from the following distribution:
\begin{equation}\label{eq:setup}
    y\stackrel{u.a.r}{\sim} \{-1, +1\},\quad
    x\stackrel{i.i.d.}{\sim}\mathcal{N}(y\alpha\phi_1+\beta\phi_2, 1).
\end{equation}
Here, $\phi_1\in\mathbb{R}^2$ and $\phi_2\in\mathbb{R}^2$ represent class-discriminative and class-common feature basis vectors, respectively, where $\phi_1^\top\phi_2=0$ and $\| \phi_1 \| = \| \phi_2 \| = 1$.
$\alpha$ and $\beta$ denote the strength of the class-discriminative and class-common features, respectively, where $\alpha \geq 1$ and $\beta \geq 0$.
We generate a reduced dataset $\mathcal{S}=\{\mathcal{S}^+,\mathcal{S}^-\}$ of $\mathcal{X}$, where $\mathcal{S}^+$ and $\mathcal{S}^-$ are $(s_1,+1)$ and $(s_2,-1)$, respectively. We use a linear classifier $f(x)=\textup{sign}(w^\top x)$ and the hinge loss function $L(x,y;w)=\max\left(0, 1-yw^\top x\right)$, where $w=\phi_1$. For convenience of description, we define $\mathcal{X}^+=\{(x_i,y_i)\mid i\in\{1,\cdots,N\},-y_iw^\top x_i<1, y_i=+1\}$ and $\mathcal{X}^-=\{(x_j,y_j)\mid j\in\{1,\cdots,N\},-y_jw^\top x_j<1, y_j=-1\}$. We define a $\ell_2\text-$distance-based gradient matching loss as follows
($\mathcal{X}$: the training dataset, $\mathcal{S}$: the synthetic dataset):
\begin{equation}\label{eq:ex_matching_loss}
\begin{gathered}
\mathcal{L}(\mathcal{X},\mathcal{S};w)= \frac{\lambda}{|\mathcal{S}|}\sum_{s\in\mathcal{S}}\|s\|\\
+\left\| \frac{1}{|\mathcal{X}|}\sum_{(x,y)\in\mathcal{X}}g_w(x,y)-\frac{1}{|\mathcal{S}|}\sum_{(s,t)\in\mathcal{S}}g_w(s,t) \right\|,
\end{gathered}
\end{equation}
where $g_w(\cdot)=\nabla_wL(\cdot;w)$.
In our example, $\lambda\in\mathbb{R}^+$ is a control parameter of the capacity of the synthetic dataset $\mathcal{S}$. Here, we assume that $\lambda$ is selected by making $\max_{s \in \mathcal S} \| s \|$ upper bounded by $\epsilon\leq1-\frac{\sqrt{2}}{\sqrt{\pi}}$. Finally, we define a class-discriminative and class-common feature ratio $R(\mathcal{S})$ to evaluate the quality of the generated $\mathcal{S}$ as follows:
\begin{equation}\label{eq:evaluate_R}
    R(\mathcal{S})=\frac{1}{|\mathcal{S}|}\sum_{s\in\mathcal{S}}\frac{\abs{s^\top\phi_1}}{\abs{ s^\top\phi_1} + \abs{s^\top\phi_2}},
\end{equation}
where, $R(\mathcal{S})=1$ indicates that $\mathcal{S}$ contains only class-discriminative features, whereas $R(\mathcal{S})=0$ indicates that $\mathcal{S}$ holds only class-common features.

\paragraph{Issues with class-wise gradient matching.}
The optimal solution of Eq. \eqref{eq:ex_matching_loss}
for the class-wise gradient matching strategy, employed by previous DC approaches \cite{dc, dsa} is as the follows:
\begin{equation} \label{eq:class_wise}
\begin{gathered}
    \tilde{\mathcal{S}}=\argmin_\mathcal{S}\mathcal{L}(\mathcal{X}^+,\mathcal{S}^+)+\mathcal{L}(\mathcal{X}^-,\mathcal{S}^-)\\
    =\argmin_\mathcal{S}\left\| \mu^+-s_1 \right\|+\left\| \mu^--s_2 \right\|+\lambda_\mathcal{S}\\
    =\left\{\left(\frac{\epsilon\mu^+}{\|\mu^+\|},+1\right), \left(\frac{\epsilon\mu^-}{\|\mu^-\|},-1\right)\right\},
\end{gathered}
\end{equation}
where $\mu^+=\frac{1}{|\mathcal{X}^+|}\sum_{x\in\mathcal{X}^+}x$, $\mu^-=\frac{1}{|\mathcal{X}^-|}\sum_{x\in\mathcal{X}^-}x$, and $\lambda_\mathcal{S}=\lambda\sum_{s\in\mathcal{S}}\|s\|$.
More detailed equations can be found in Appendix~B.
Equation~\eqref{eq:class_wise} demonstrates that the class-wise gradient matching method optimizes $\mathcal{S}$ for each class to ensure that it has the same direction as the average of the training samples that generate gradients.
Then, $R(\tilde{\mathcal{S}})$ is:
\begin{equation}\label{eq:R_class_wise}
R\left( \tilde{\mathcal{S}} \right)\leq\frac{\alpha}{\alpha+\beta}.
\end{equation}
The equality holds when $\beta=0$, and the inequality is due to $\abs{\frac{\phi_1^\top\mu^+}{\| \mu^+ \|}}=\abs{\frac{\phi_1^\top\mu^-}{\| \mu^- \|}}<\alpha$.
Equation~\eqref{eq:R_class_wise} shows that when $\alpha \ll \beta$ (\ie, class-common features are dominant and class-discriminative features are minority), $R(\tilde S) \rightarrow 0$, that is, the class-wise gradient matching method can result in synthetic datasets that are ineffective for the classification task.
For example, as shown in Table \ref{tab:finegrained}, the class-wise gradient matching method can fail when applied to fine-grained classification tasks that include shared appearance between classes and can be discriminated only by fine-grained appearances.

\paragraph{Leveraging contrastive signals.}
The class-wise gradient matching method has a limitation when class-common features are dominant. We need a different approach to capture only class-discriminative features for better downstream task performance. The following simple modification of Eq. \eqref{eq:class_wise} can mitigate this issue:
\begin{equation} \label{eq:class_comprehensive}
    \begin{gathered}
    \hat{\mathcal{S}}=\argmin_\mathcal{S}\mathcal{L}(\mathcal{X}^+\cup\mathcal{X}^-,\mathcal{S}^+\cup\mathcal{S}^-)\\
    =\argmin_\mathcal{S}\left\| (\mu^+-\mu^-)-(s_1-s_2) \right\|+\lambda_\mathcal{S} \\
    =\left\{\left(\epsilon\phi_1,+1\right), \left(-\epsilon\phi_1,-1\right)\right\}.
\end{gathered}
\end{equation}
Equation~\eqref{eq:class_comprehensive} considers the loss gradients for all classes collectively, whereas Eq. \eqref{eq:class_wise} considers the loss gradients for each class separately. Moreover, Eq. \eqref{eq:class_comprehensive} reveals that the sum of loss gradients between classes is important because it contains contrastive signals between classes ($(\mu^+-\mu^-)$ and $(s_1-s_2)$).
Here, $R(\hat{\mathcal{S}})$ is calculated as follows:
\begin{equation}\label{eq:R_class_collective}
R\left( \hat{\mathcal{S}} \right)=1.
\end{equation}
In other words, $\hat{\mathcal{S}}$ contains only class-discriminative features to ensure that it is independent of the proportion of class-common features in the original training dataset $\mathcal{X}$.

\paragraph{Empirical evidence.}
Here, we empirically demonstrate that the arguments developed above, based on a simple theoretical model, can also be applied to modern machine learning settings. To be specific, we (\romannumeral 1) define a binary classification task (3 vs. 8) using MNIST; (\romannumeral 2) train a convolutional neural network (CNN) model on the binary task using the cross-entropy loss; (\romannumeral 3) generate reduced datasets of the task (3 vs. 8) by applying the class-wise gradient matching method (DC) and the class-collective gradient matching method (Eq. \eqref{eq:dcc}), respectively; and (\romannumeral 4) horizontally flip all training images from the class ``3'' and repeat (\romannumeral 2) to (\romannumeral 3).
Digits ``3'' and ``8'' can be easily classified by the difference in shape on the left halves (discriminative features), while the right halves look almost identical (common features).

\input{Figure/figure_empirical_evidence}
Figure~\ref{fig:empirical_evidence} illustrates images synthesized by the DC and our proposed method. The figure shows that
the class-wise gradient matching method generates near-prototype images for each class. In contrast, the class-collective gradient matching method optimizes synthetic images by prioritizing the difference between the two classes. 
For example, the red boxes in Fig. \ref{fig:empirical_evidence_b} show that our class-collective gradient matching method synthesizes the images of class ``8'' with an emphasis on the left half. The same trend can be found in Fig. \ref{fig:empirical_evidence_c}, indicating that the results are not due to chance or dataset bias, but because the class-collective method leverages contrastive signals.
For simple tasks such as MNIST, however, our motivation may not lead to improvements compared to DC, because the number of features in the training dataset is limited to ensure the efficiency of the condensation method.
However, for complex tasks that need to capture subtle differences between classes, our approach can result in significant improvements in dataset condensation.

\subsection{Dataset Condensation with Contrastive Signals}\label{subsec:dcc}
Based on Sec. \ref{subsec:motivating_example}, we propose Dataset Condensation with Contrastive signals (DCC). The DCC optimizes a synthetic dataset by minimizing the following objective function:
\begin{equation} \label{eq:dcc}
\begin{gathered}
    \mathop{\mathbb{E}}_{\theta_0\sim P_{\theta_0}}\left[ \sum_{t=0}^{T-1}D\left( 
    \sum_{c=0}^{C-1}g_{\theta_t}(\mathcal{X}^c), \sum_{c=0}^{C-1}g_{\theta_t}(\mathcal{S}^c)
    \right) \right],\\
    \textup{subject to}\: \theta_{t+1}=\theta_t-\frac{\eta}{|\mathcal{S}|}\sum_{(s,t)\in\mathcal{S}}\nabla_{\theta_t}L(s,t;\theta_t).
\end{gathered}
\end{equation}
Here, $g_{\theta_t}(\mathcal{X}^c)=\frac{1}{|\mathcal{X}^c|}\sum_{(x,y)\in\mathcal{X}^c}\nabla_{\theta_t} L(x,y;\theta_t)$, where $\mathcal{X}^c=\{(x,y)\mid(x,y)\in\mathcal{X}, y=c\}$. $D(\cdot,\cdot)$ and  $L(\cdot;)$ denote the distance function and cross-entropy loss function, respectively.
We find the solution to Eq. \eqref{eq:dcc} by alternately training the network parameters $\theta_t$ and synthetic dataset $\mathcal{S}$, with the periodic initialization of the classifier as in DC. We name the loops initializing $\theta$ and updating $\mathcal{S}$ ``outer-loop'' and ``inner-loop,'' respectively.
The primary difference between Eq. \eqref{eq:dcc} and the objective functions of existing methods (Eq. \eqref{eq:dc}) are the locations of the summation over classes $\sum_{c=0}^{C-1}$. Existing methods first determine the gradient distance for each class and then sum them up, while DCC sums up the gradients over the classes first and then measures the gradient distance between the training and synthetic datasets. Therefore, as implied in Sec. $\ref{subsec:motivating_example}$, DCC can effectively leverage the contrastive signals present in the sum of loss gradients over classes, thereby synthesizing small datasets that are more suitable for classification tasks.

\paragraph{A bi-level warm-up strategy.}\label{sec:bilevel_warmup}
\input{Figure/figure_kernel_velocity}
DNNs are known to undergo chaotic transience during the early phase of training \cite{kernel_velocity,gsnr}.
In addition, during the dataset condensation process, the classifier is periodically initialized, as described in Sec. \ref{subsec:dc}, thereby repeatedly inducing the chaotic training phase of the classifier.
We analyze the impact of this periodic transience on the training dynamics of the DCC by measuring the Neural Tangent Kernel (NTK) velocity \cite{kernel_velocity} on the synthetic dataset. \citet{kernel_velocity} introduced the NTK velocity to characterize the loss landscape geometry and training dynamics of DNNs. The NTK velocity is the time evolution of the data-dependent NTK, which, in our case, is the Gram matrix of the Jacobian of the gradient matching loss with respect to the synthetic data samples. The high NTK velocity indicates that the loss landscape is highly nonlinear, and thus, the update direction of the synthetic dataset changes rapidly. 

\input{Figure/algorithm_v2}
Figure~\ref{fig:kernel_velocity} shows the NTK velocity during synthetic dataset optimization using DC and DCC on CIFAR-10. As shown, the NTK velocity periodically repeats the process of peaking at the classifier initialization and then rapidly stabilizes. Moreover, the peaks of DCC are much higher than those of DC. This difference is reasonable, because when synthesizing images with the class label ``$c$,'' DC obtains the loss gradient using only the training data of the class ``$c$,'' while DCC obtains the loss gradient from all the classes. Thus, noisy gradients from the other classes can be excluded in DC, whereas DCC may accumulate noisy gradients from all the classes. Although the higher peaks are not detrimental in terms of optimization \cite{break-even_point,kernel_velocity}, we empirically determine that the peaks during the early phase of dataset condensation can suppress the effectiveness of DCC (see Table \ref{tab:ablation}).

To address this issue, we introduce \emph{a bi-level warm-up strategy} for the DCC. We define the inner-loop level (updating $\mathcal{S}$) and outer-loop level (initializing $\theta$) warm-up and apply class-wise gradient matching under the two warm-up conditions. The overall procedure for the proposed method is described in Algorithm~\ref{algo}.

%% file: Figure/figure_motivating_example.tex
\begin{figure}[ht]
{
\begin{center}
\centerline{\includegraphics[width=.9\columnwidth]{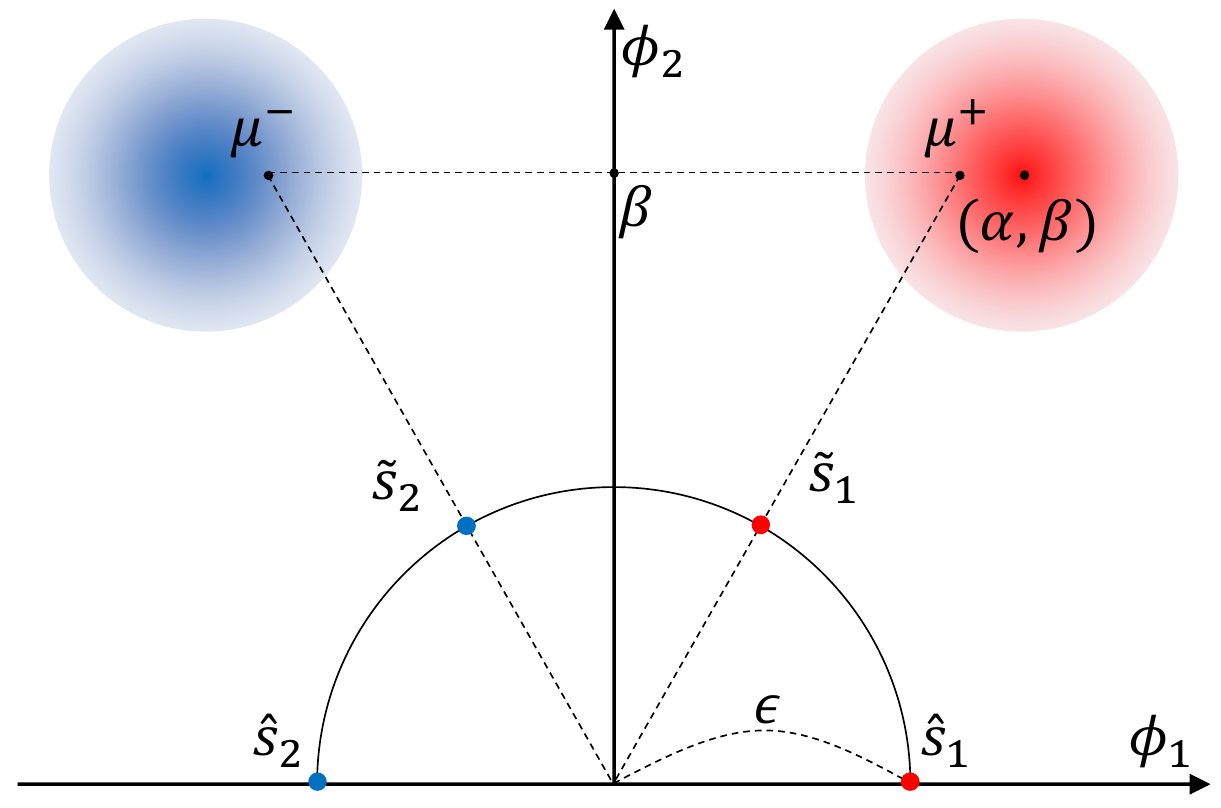}}
\caption{Overview of Sec. \ref{subsec:motivating_example}. Red and blue circles denote the data distributions of $y=+1$ and $-1$, respectively.}
\label{fig:motivating_example}
\end{center}
}
\end{figure}

%% file: Figure/figure_empirical_evidence.tex
\begin{figure}
{
\centering
\begin{subfigure}{\columnwidth}
    \includegraphics[width=\columnwidth]{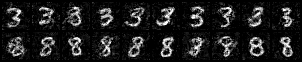}
    \vskip -0.07in
    \caption{3 vs. 8, class-wise gradient matching.}
    \label{fig:empirical_evidence_a}
\end{subfigure}
\begin{subfigure}{\columnwidth}
    \includegraphics[width=\columnwidth]{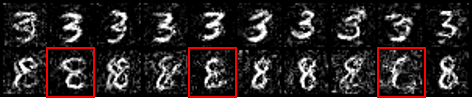}
    \vskip -0.07in
    \caption{3 vs. 8, class-collective gradient matching (ours).}
    \label{fig:empirical_evidence_b}
\end{subfigure}
\begin{subfigure}{\columnwidth}
    \includegraphics[width=\columnwidth]{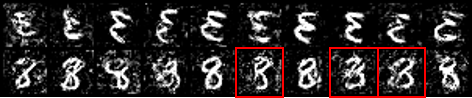}
    \vskip -0.07in
    \caption{Flipped 3 vs. 8, class-collective gradient matching (ours).}
    \label{fig:empirical_evidence_c}
\end{subfigure}
        
\caption{Generated images (10 images per class) for each setting (shown below each subfigure). We mark the images we want to emphasize with red boxes.}
\label{fig:empirical_evidence}
}
\end{figure}

%% file: Figure/figure_kernel_velocity.tex
\begin{figure}[ht]
{
\begin{center}
\centerline{\includegraphics[width=\columnwidth]{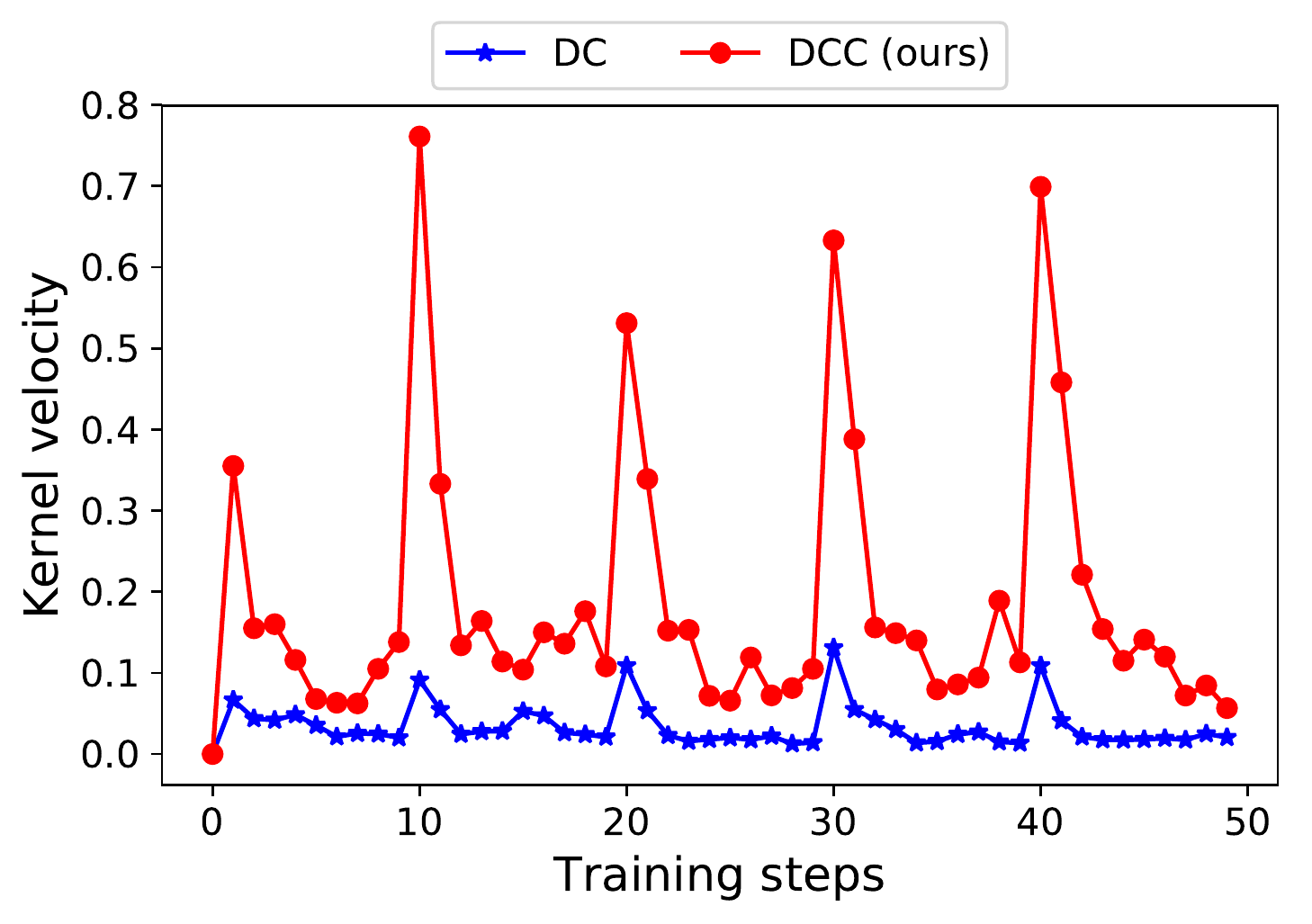}}
\caption{NTK velocity during the synthetic dataset optimization using DC and DCC on CIFAR-10.}
\label{fig:kernel_velocity}
\end{center}
}
\end{figure}

%% file: Figure/algorithm_v2.tex
\renewcommand{\algorithmiccomment}[1]{#1}
\begin{algorithm}[h]
\caption{Dataset condensation with contrastive signals}
\begin{algorithmic}[1]
\label{algo}
\REQUIRE Training datset $\mathcal{X}$, synthetic dataset $\mathcal{S}$, outer/inner-loop iterations $K_o, K_i$, network training iterations $T$, outer/inner-loop level warm-up iterations $\gamma_o, \gamma_i$, learning rate for synthetic images and network $\tau$, $\eta$, number of images per class $\zeta$
\STATE Initialize $\mathcal{S}$ with a subset of $\mathcal{X}$ s.t. $|S^c|=\zeta$, $\forall$ class $c$
\FOR[{ \textcolor{blue}{\;\# outer-loop}}]{$k_o=0$ \textbf{to} $K_o-1$}
\STATE Initialize the network parameter $\theta$
\STATE \texttt{warmup}, $k_i$ $\leftarrow$ True, 0
\WHILE[{ \textcolor{blue}{\;\# inner-loop with warm-up}}]{\texttt{warmup}}
\IF[{ \textcolor{blue}{\;\# bi-level warm-up}}]{$k_o>\gamma_o$ \textbf{or} $k_i>\gamma_i$}
\STATE \texttt{break};
\ENDIF
\STATE \textcolor{blue}{\# class-wise gradient matching loss}
\STATE Compute $\mathcal{L}$ by Eq. \eqref{eq:dc}
\STATE $\mathcal{S}\leftarrow \mathcal{S}-\tau\cdot\nabla_{\mathcal{S}}\mathcal{L}$\textcolor{blue}{\;\;\# synthetic images update}
\STATE Update $\theta$ using $\mathcal{S}$ for $T$ iterations
\STATE $k_i \leftarrow k_i + 1$
\ENDWHILE
\STATE $g_\mathcal{X}$, $g_\mathcal{S}$ $\leftarrow$ 0, 0
\WHILE[{ \textcolor{blue}{\;\# inner-loop without warm-up}}]{$k_i < K_i$}
\FOR{$c=0$ \textbf{to} $C-1$}
\STATE Sample a minibatch pair $\bar{\mathcal{X}}^c\sim\mathcal{X}$ and $\bar{\mathcal{S}}^c\sim\mathcal{S}$
\STATE $g_\mathcal{X}, g_\mathcal{S} \leftarrow g_\mathcal{X}+g_{\theta}(\bar{\mathcal{X}}^c), g_\mathcal{S}+g_{\theta}(\bar{\mathcal{S}}^c)$
\ENDFOR
\STATE \textcolor{blue}{\# class-collective gradient matching loss}
\STATE $\mathcal{L} \leftarrow D(g_\mathcal{X}, g_\mathcal{S})$
\STATE $\mathcal{S}\leftarrow \mathcal{S}-\tau\cdot\nabla_{\mathcal{S}}\mathcal{L}$\textcolor{blue}{\;\;\# synthetic images update}
\STATE Update $\theta$ using $\mathcal{S}$ for $T$ iterations
\STATE $k_i \leftarrow k_i + 1$
\ENDWHILE
\ENDFOR
\STATE \textbf{Output:} a synthetic dataset $\mathcal{S}$
\end{algorithmic}
\end{algorithm}

%% file: experiments.tex
\section{Experimental Results and Discussion}
\subsection{Experimental Setup}
\paragraph{Datasets.}
We complement our analysis with experiments conducted on SVHN, CIFAR-10, CIFAR-100, and the fine-grained image classification datasets (Automobile, Terrier, Fish, Truck, Insect, and Lizard) subsampled from ImageNet32x32 \cite{downsampled_imagenet} using the WordNet hierarchy \cite{wordnet}. A detailed description of the datasets is summarized in Appendix~D.

\paragraph{Implementation Details.}
In our experiments, we compare the proposed method with the baseline methods for the settings of learning 1, 10, and 50 image(s) per class as in \citet{dc}. We use ConvNet \cite{convnet} as a classifier from which the gradients for matching are obtained in the dataset condensation process. We set $K_o=1000$, $\gamma_o=250$, $\gamma_i=10$, and $\tau=0.1$. For settings of learning 1, 10, and 50 image(s) per class, ($K_i, T$) is set to (10,5), (10,50), and (50,10), respectively. To reduce the training data based on the selection-based methods, we use the pre-computed scores provided by \citet{cscores} (C-scores) and the average of the scores computed for 10 independently pre-trained models (GraNd and EL2N). To evaluate the selection-based methods, we train 100 classifiers on the coreset from scratch and obtain the mean and standard deviation of their test accuracies. In addition, to evaluate the synthesis-based methods, including our proposed method, we learn five synthetic datasets and train 20 classifiers from scratch on each synthetic dataset to obtain the mean and standard deviation of 100 test accuracies. Please refer to \cite{dc,dsa} for more details. For the implementation of KIP \citep{kip}, we use the code\footnote{\url{https://colab.research.google.com/github/google-research/google-research/blob/master/kip/KIP.ipynb}} provided by the authors. Note that we denote the DCC with differentiable Siamese augmentation \citep{dsa} as DSAC.

\subsection{Dataset Condensation}
\paragraph{Results on Fine-Grained Datasets.}
\input{Table/table_finegrained}
\input{Figure/figure_imagenet_automobile}
\input{Table/table_benchmark}
We first evaluate the improvements over the baselines of the proposed method on fine-grained image classification datasets. Tab.~\ref{tab:finegrained} shows that the results of DC are consistent with those of the motivation example described in Sec.~\ref{subsec:motivating_example}. In particular, DC perform worse than random selection (Random) on the Automobile, Terrier, and Fish datasets. In contrast, DCC always performs better than Random, implying that the proposed method effectively considers the class-discriminative features. Moreover, DSAC always outperforms DSA, showing that improved methods using diverse image transformations do not effectively detect differences between classes. In addition, although the proposed method is largely orthogonal to differentiable Siamese augmentation, it can be observed that DCC is more effective than DSAC in certain cases.
Considering image transformations as a form of regularization \cite{da_reg}, we hypothesize that such cases indicate that applying additional regularization may hinder the optimization process of the proposed method.
Finally, we qualitatively compare the synthetic images for the classes of Automobile generated through each method of DCC and DC. Figure~\ref{fig:visual_automobile} shows that the images learned by the proposed method are sharper than those learned by DC and display the unique patterns of each class more prominently. For example, the long body and multiple windows of the ``limousine'' or the distinctive body frame of ``Model T'' are clearly exhibited in the results of DCC (red box in Fig.~\ref{fig:visual_automobile}), while the differences between the classes are ambiguous and difficult to distinguish in the results of DC.

\paragraph{Results on Benchmark Datasets.}
\input{Figure/figure_alignment_uniformity}
Table~\ref{tab:benchmark} presents the comparison results of selection-based methods, synthesis-based methods, and our methods on SVHN, CIFAR-10 and CIFAR-100.
First, in contrast to the observation by \citet{dc}, recent selection-based methods achieve better results than Random for the settings of 1 and 10 image(s) per class. However, their performance is still worse than that of the synthesis-based methods with large gaps.
Although KIP performs well in some settings, it can be observed that its effectiveness is unstable depending on the dataset or Img/cls. Unlike fine-grained classification tasks, DC and DSA always show improvements compared to Random and the selection-based methods with large gaps (\eg, +15\% in SVHN).
Nevertheless, we observe that our method achieves the best performance not only for fine-grained tasks, but also for general vision classification benchmarks.

To understand the significant improvements achieved by our method, we provide an additional analysis of our method and DC from the perspective of representation learning. \citet{wang2020uniformity} recently demonstrated that two metrics, \ie, uniformity (how features are uniformly distributed on the feature space) and alignment (how two features with the same class are close) are highly correlated to the quality of the learned representations. Following \citet{wang2020uniformity}, we plot the uniformity and alignment losses for the features of DC and DCC in Fig. \ref{fig:alignment_uniformity}. We extract the features using ResNet-18 \cite{resnet} and VGG-11 \cite{vgg}, which are pre-trained on CIFAR-10.
In the figure, DCC shows lower uniformity loss (\ie, samples are more uniformly distributed) and lower alignment loss (\ie, positive samples are closer) than DC. That is, as in our observations, DCC can capture task-relevant features that help classification tasks by using contrastive signals.


\paragraph{Cross-Architecture Generalization.}
\input{Table/table_crossarch}
We test the generalizability of the synthetic dataset learned through the the proposed method. In particular, we train various CNN architectures, including ConvNet, LeNet \cite{lenet}, AlexNet \cite{alexnet}, VGG-11, and ResNet-18, on small datasets created using Random, DC, DSA, DCC, and DSAC and list their mean test accuracy in Tab.~\ref{tab:crossarch}. As shown, our proposed method achieves improved results not only for the architecture used for condensation, but also for other CNN architectures tested.

\paragraph{Ablation Study.}
We demonstrate the importance of the bi-level warm-up strategy when applying the proposed method. We compare the performance of the proposed method with and without the bi-level warm-up strategy and present the results in Tab.~\ref{tab:ablation}. In the table, \textcolor{red}{\ding{55}} is equivalent to Algorithm~\ref{algo} where $\gamma_o=0$ and $\gamma_i=0$, while \textcolor{green}{\ding{51}} indicates $\gamma_o=250$ and $\gamma_i=10$. From the results, it can be seen that when the capacity of the synthetic dataset is relatively large, the bi-level warm-up has a negligible effect, whereas when the budget is limited, the bi-level warm-up affects significant performance improvements. In particular, without the bi-level warm-up strategy, DCC yields worse results than the baseline methods (DC and DSA) for settings of learning 1 and 10 images per class of CIFAR-100, highlighting the importance of the bi-level warm-up strategy for small budgets. In addition, Tab.~\ref{tab:design_warmup} shows the ablation study on designing warm-up: None (no warm-up), Simple (warm-up dependent only on inner-loops), Proposed (bi-level). As shown, the NTK velocity peaks during the later phase of dataset condensation do not suppress the effectiveness of DCC.\input{Table/table_ablation}\input{Table/table_design_warmup}

\subsection{Application: Continual Learning}
\input{Figure/figure_continual}
We apply our method to a continual learning task, where the training datasets are sequentially input
with task labels.
We build our method on a popular memory-based continual learning baseline, called Experience Replay with Ring Buffer strategy (ER-RB) \cite{chaudhry2019tiny}. This baseline randomly stores the same amount of data per class of old tasks and replays them to avoid forgetting old tasks while learning a new task.
To observe the effectiveness of the applications of the dataset condensation methods on continual learning tasks, we substitute DSA and DSAC for the ring buffer strategy.
We train models on a sequence of three fine-grained image datasets (\textit{i.e.}, \{Lizard-Truck-Insect\}) with the ER-RB, DSA, and DSAC methods (10 images per class) and compare them in terms of the average accuracy of seen tasks in Figure \ref{fig:continual}.
As shown, DSAC outperforms RB and DSA by 6.7\% and 2.5\% in $T3$, respectively.
These results indicate that the dataset generated by DSAC is more informative than those created by other baselines and hence more helpful in preventing memory loss of past tasks.
We provide the results of the continual learning task on general vision classification benchmarks in Appendix~C.

%% file: Table/table_finegrained.tex
\newcommand{\rednum}[1]{\textbf{\textcolor{red}{#1}}}
\newcommand{\bluenum}[1]{\textcolor{blue}{#1}}

\begin{table}[]
\centering
\caption{
Comparison of the proposed method with the baselines (Random, DC and DSA) on fine-grained image classification datasets. Each number is the average over 100 different runs.
The \bluenum{blue} and \rednum{red} numbers denote worse than Random and the best results, respectively.} \label{tab:finegrained}

{\footnotesize
\addtolength{\tabcolsep}{-2pt}
\begin{tabular}{ccccccc}
\toprule
\multirow{2}{*}{Dataset} & \multirow{2}{*}{Img/cls} & \multirow{2}{*}{Random} & \multicolumn{2}{c}{Baselines} & \multicolumn{2}{c}{Ours} \\
& & & DC & DSA & DCC & DSAC \\\midrule

\multirow{2}{*}{Automobile} & 10 & 12.2 & \bluenum{11.0} & 19.1 & 18.6 & \rednum{22.1} \\
& 50 & 19.5 & \bluenum{16.8} & 24.1 & 28.3 & \rednum{29.2} \\\midrule

\multirow{2}{*}{Terrier} & 10 & 5.6 & \bluenum{4.6} & \bluenum{5.1} & \rednum{6.4} & 6.2 \\
& 50 & 7.8 & \bluenum{4.8} & \bluenum{7.2} & 10.7 & \rednum{10.7} \\\midrule

\multirow{2}{*}{Fish} & 10 & 14.7 & \bluenum{13.5} & 18.7 & 20.4 & \rednum{22.3} \\
& 50 & 15.3 & 17.0 & 19.4 & \rednum{28.4} & 23.3 \\\midrule

\multirow{2}{*}{Lizard} & 10 & 13.3 & 23.5 & 29.1 & 30.0 & \rednum{34.2} \\
& 50 & 20.9 & 32.6 & 32.9 & \rednum{38.8} & 34.8 \\\midrule

\multirow{2}{*}{Truck} & 10 & 21.2 & 24.8 & 36.5 & 39.4 & \rednum{48.1} \\
& 50 & 31.8 & 43.5 & 57.9 & 57.4 & \rednum{60.6} \\\midrule

\multirow{2}{*}{Insect} & 10 & 27.6 & 41.8 & 47.4 & 48.7 & \rednum{50.0} \\
& 50 & 42.7 & 49.6 & 51.3 & \rednum{55.8} & 51.9 \\\bottomrule

\end{tabular}
}

\end{table}







%% file: Figure/figure_imagenet_automobile.tex
\begin{figure}[ht]
{
\begin{subfigure}{.495\columnwidth}
  \centering
  \includegraphics[width=\columnwidth]{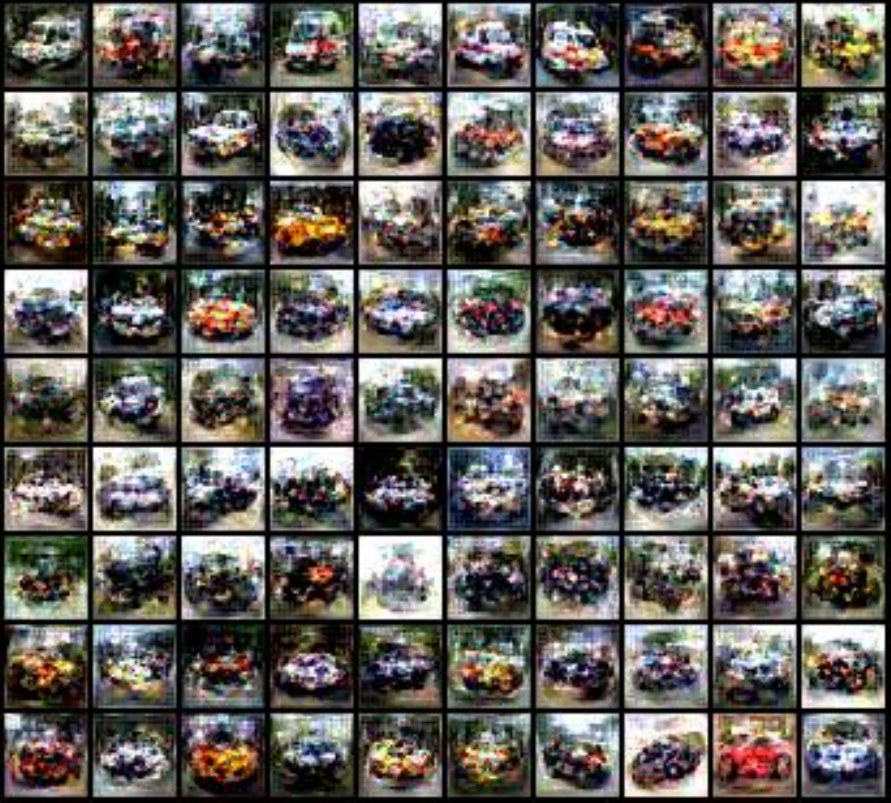}  
  \caption{DC}
  \label{fig:visual_automobile_dc}
\end{subfigure}
\begin{subfigure}{.495\columnwidth}
  \centering
  \includegraphics[width=\columnwidth]{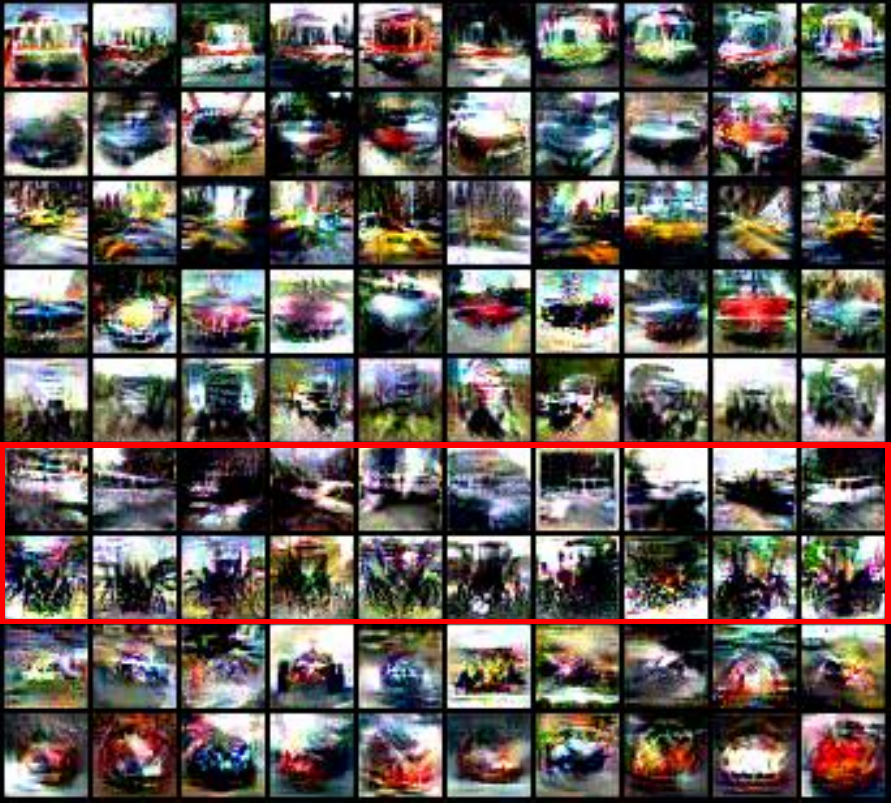}  
  \caption{DCC}
  \label{fig:visual_automobile_dcc}
\end{subfigure}
\caption{Visualization of the generated 10 images per class of Automobile. From the top row, ``ambulance'', ``beach wagon'', ``cab'', ``convertible'', ``jeep'', ``limousine'', ``Model T'', ``racer'', and ``sports car''.}
\label{fig:visual_automobile}
}
\end{figure}

%% file: Table/table_benchmark.tex
\begin{table*}[]
\centering
\caption{Comparison of the performance (mean$\pm$std \%) of the proposed method with the selection-based (Random, C-score, GraNd, and EL2N) and sysnthesis-based (KIP, DC, and DSA) methods on benchmark datasets.
Img/cls stands for the number of images per class, and $\Omega$ denotes the upper bound of the performance, which can be obtained by learning the original training dataset.
The best results within each setting (Dataset, Img/cls) are indicated in bold.} \label{tab:benchmark}

{\footnotesize
\addtolength{\tabcolsep}{-3pt}
\begin{tabular*}{\textwidth}{cccccccccccc}
\toprule
\multirow{2}{*}{Dataset} & \multirow{2}{*}{Img/cls} & \multicolumn{4}{c}{Selection-based} & \multicolumn{3}{c}{Synthesis-based} & \multicolumn{2}{c}{Ours} & \multirow{2}{*}{$\Omega$} \\
&& Random & C-score & GraNd & EL2N & KIP & DC & DSA & DCC & DSAC & \\\midrule

\multirow{3}{*}{SVHN} & 1 & 14.6$\pm$1.6 & - & 19.6$\pm$0.5 & 19.1$\pm$0.6 & 23.3$\pm$2.7& 34.6$\pm$2.0 & 36.0$\pm$2.0 & 34.3$\pm$1.6 & \textbf{47.5$\pm$2.6} & \multirow{3}{*}{92.1$\pm$0.2}\\

& 10 & 35.1$\pm$4.1 & - & 37.5$\pm$1.6 & 32.5$\pm$1.2 & 62.4$\pm$0.5& 76.2$\pm$0.6 & 78.9$\pm$0.5 & 76.2$\pm$0.8 & \textbf{80.5$\pm$0.6} &\\

& 50 & 70.9$\pm$0.9 & - & 69.1$\pm$0.7 & 68.7$\pm$0.7 & 69.6$\pm$0.5& 82.7$\pm$0.3 & 84.4$\pm$0.4 & 83.3$\pm$0.2 & \textbf{87.2$\pm$0.3} &\\\midrule

\multirow{3}{*}{CIFAR-10} & 1 & 14.4$\pm$2.0 & 21.7$\pm$0.6 & 21.8$\pm$0.5 & 20.9$\pm$0.6 & \textbf{37.6$\pm$1.0}& 28.2$\pm$0.7 & 28.7$\pm$0.7 & 32.9$\pm$0.8 & 34.0$\pm$0.7 & \multirow{3}{*}{81.6$\pm$0.3}\\

& 10 & 26.0$\pm$1.2 & 31.6$\pm$0.4 & 32.3$\pm$0.4 & 32.3$\pm$0.4 & 47.3$\pm$0.3& 44.7$\pm$0.6 & 52.1$\pm$0.6 & 49.4$\pm$0.5 & \textbf{54.5$\pm$0.5} &\\

& 50 & 43.4$\pm$1.0 & 39.8$\pm$0.4 & 41.2$\pm$0.3 & 40.7$\pm$0.3 & 50.1$\pm$0.2& 54.8$\pm$0.5 & 60.6$\pm$0.4 & 61.6$\pm$0.4 & \textbf{64.2$\pm$0.4} &\\\midrule

\multirow{3}{*}{CIFAR-100} & 1 & 4.2$\pm$0.3 & 8.0$\pm$0.3 & 8.8$\pm$0.3 & 8.8$\pm$0.3 & \textbf{14.8$\pm$1.2}& 12.8$\pm$0.3 & 13.9$\pm$0.4 & 13.3$\pm$0.3 & 14.6$\pm$0.3 & \multirow{3}{*}{52.5$\pm$0.3}\\

& 10 & 14.6$\pm$0.5 & 18.1$\pm$0.2 & 17.8$\pm$0.2 & 17.3$\pm$0.2 & 13.4$\pm$0.2& 26.6$\pm$0.3 & 32.4$\pm$0.3 & 30.6$\pm$0.4 & \textbf{33.5$\pm$0.3} &\\

& 50 & 29.7$\pm$0.4 & 30.4$\pm$0.3 & 27.6$\pm$0.2 & 27.7$\pm$0.2 & - & 32.1$\pm$0.3 & 38.6$\pm$0.3 & \textbf{40.0$\pm$0.3} & 39.3$\pm$0.4 &\\\bottomrule

\end{tabular*}}

\end{table*}

%% file: Figure/figure_alignment_uniformity.tex
\begin{figure}[ht]
{
\begin{center}
\centerline{\includegraphics[width=\columnwidth]{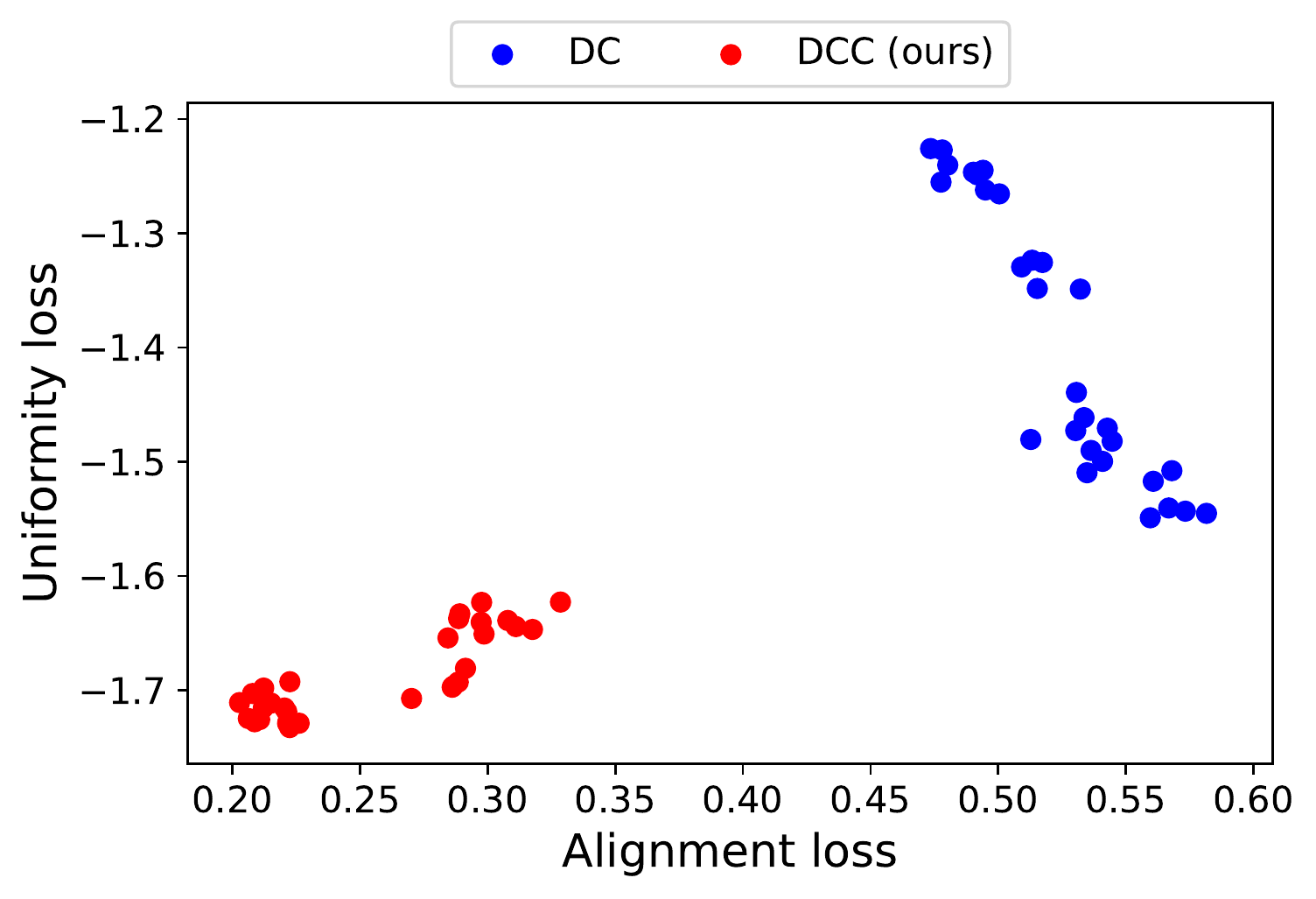}}
\caption{Alignment and uniformity loss for the features of synthetic images generated by DC and DCC on CIFAR-10 (50 images per class). We used 6 pre-trained networks and 10 synthetic datasets (5 DC + 5 DCC), which result in total 60 points in this plot.}
\label{fig:alignment_uniformity}
\end{center}
}
\end{figure}

%% file: Table/table_crossarch.tex
\begin{table}[]
\centering
\caption{Comparison of the cross-architecture generalization performance (the mean test accuracy over 100 runs) of the proposed method with the baseline methods using ConvNet as the source network on CIFAR-10 (50 images per class). The best results for each target network are shown in bold.} \label{tab:crossarch}

{\footnotesize
\addtolength{\tabcolsep}{-2pt}
\begin{tabular}{cccccc}
\toprule
\multirow{2}{*}{Method} & \multicolumn{5}{c}{Target network} \\
& ConvNet & LeNet & AlexNet & VGG & ResNet \\\midrule
Random & $43.2$ & $30.8$ & $35.9$ & $36.8$ & $26.1$ \\
DC & $54.8$ & $33.8$ & $40.9$ & $39.3$ & $23.9$ \\
DSA & $60.4$ & $40.3$ & $46.0$ & $50.7$ & $49.7$ \\
DCC (ours)& $61.6$ & $38.0$ & $45.2$ & $46.3$ & $27.3$ \\
DSAC (ours)& $\bm{64.1}$ & $\bm{42.6}$ & $\bm{48.2}$ & $\bm{56.0}$ & $\bm{53.9}$ \\\bottomrule

\end{tabular}
}

\end{table}

%% file: Table/table_ablation.tex
\begin{table}[]
\centering
\caption{Improvement in the effectiveness (the mean of test accuracies over 100 runs) of the proposed method on the CIFAR datasets following the application of the bi-level warm-up strategy.} \label{tab:ablation}
{\footnotesize
\addtolength{\tabcolsep}{-1pt}
\begin{tabular}{cccccc}
\toprule
\multirow{2}{*}{Dataset} & \multirow{2}{*}{Method} & \multirow{2}{*}{Bi-level warm-up} & \multicolumn{3}{c}{Img/cls} \\
& & & 1 & 10 & 50 \\\midrule
\multirow{4}{*}{CIFAR-10}& \multirow{2}{*}{DCC} & \textcolor{red}{\ding{55}} & 28.3 & 49.2 & 61.3 \\
& & \textcolor{green}{\ding{51}} & 32.9 & 49.4 & 61.6 \\\cmidrule{2-6}
& \multirow{2}{*}{DSAC} & \textcolor{red}{\ding{55}} & 32.3 & 54.0 & 63.9 \\
& & \textcolor{green}{\ding{51}} & 34.0 & 54.5 & 64.2 \\\midrule

\multirow{4}{*}{CIFAR-100}& \multirow{2}{*}{DCC} & \textcolor{red}{\ding{55}} & 12.0 & 28.5 & 40.5 \\
& & \textcolor{green}{\ding{51}} & 13.3 & 30.6 & 40.0 \\\cmidrule{2-6}
& \multirow{2}{*}{DSAC} & \textcolor{red}{\ding{55}} & 12.9 & 29.3 & 37.8 \\
& & \textcolor{green}{\ding{51}} & 14.6 & 33.5 & 39.3 \\\bottomrule

\end{tabular}}

\end{table}

%% file: Table/table_design_warmup.tex
\begin{table}[]
\centering
\caption{The results of ablation study on designing warm-up} \label{tab:design_warmup}
{\footnotesize
\addtolength{\tabcolsep}{-2pt}
\begin{tabular}{cccccc}
\toprule
\multirow{1}{*}{Dataset} & \multirow{1}{*}{Method} & \multirow{1}{*}{Img/cls} & None & Simple & Proposed \\\midrule
\multirow{4}{*}{CIFAR-100}& \multirow{2}{*}{DCC}
& 1 & 11.98 & 13.01& \textbf{13.27}\\
&& 10& 28.46 & 29.74& \textbf{30.59} \\
& \multirow{2}{*}{DSAC} 
& 1 & 12.91 & 14.21& \textbf{14.57}\\
&& 10 & 29.27 & 31.97& \textbf{33.47}\\\bottomrule
\end{tabular}
}
\end{table}

%% file: Figure/figure_continual.tex
\begin{figure}[ht]
{
\begin{center}
\centerline{\includegraphics[width=\columnwidth]{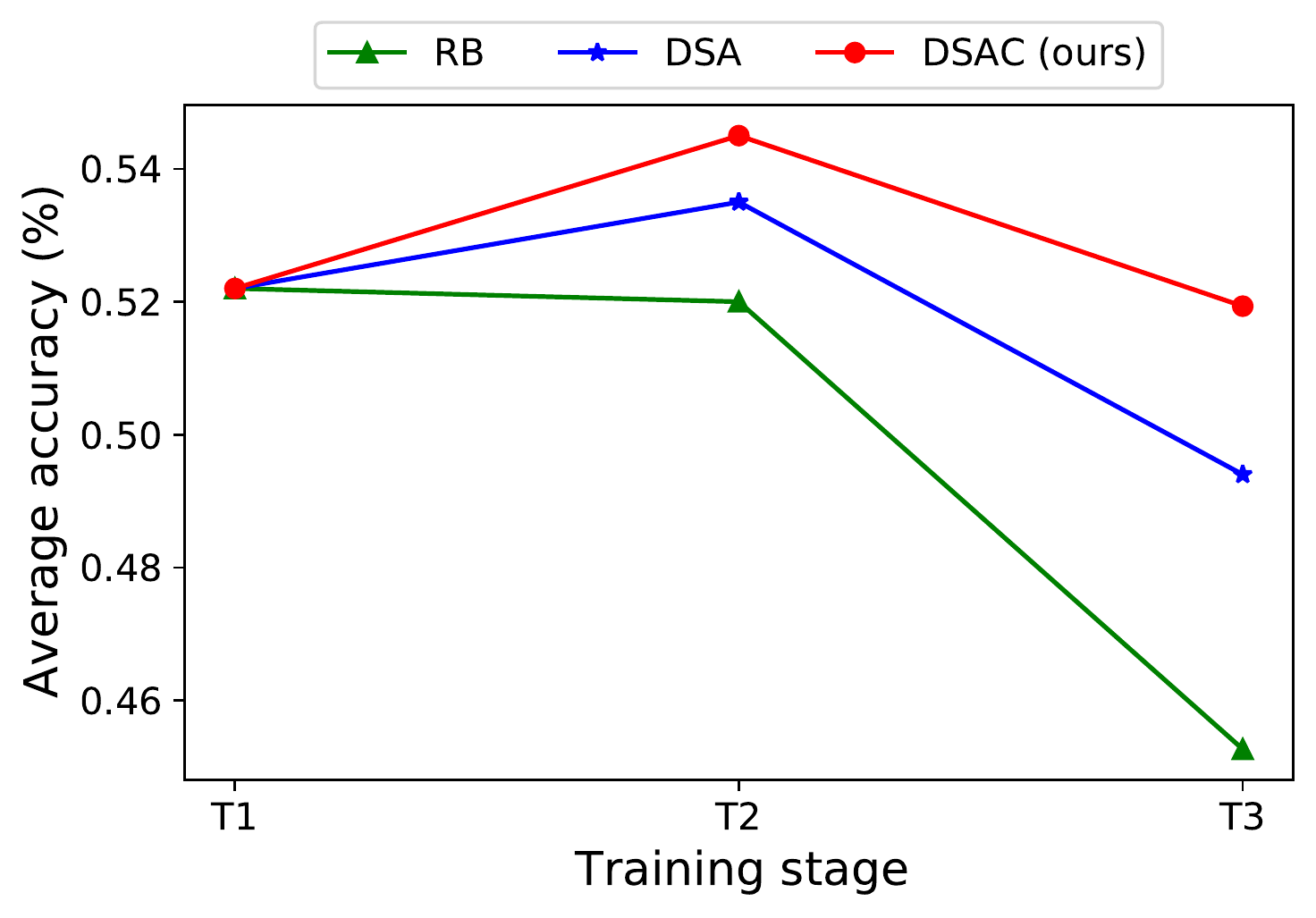}}
\caption{Performance improvements (average accuracy \%) on the continual learning task for a sequence of three fined-grained image datasets \{Lizard-Truck-Insect\} following the application of the proposed method.}
\label{fig:continual}
\end{center}
}
\end{figure}

%% file: conclusion.tex
\section{Conclusion and Future Directions}
In this study, we demonstrate that the existing dataset condensation methods perform poorly on fine-grained tasks owing to their bias toward reconstructing the prototype of each class. 
Based on the example providing the motivation for the study, we propose the DCC method, which can effectively capture subtle differences between classes through the application of class-collective gradient matching.
In addition, inspired by the training dynamics of the DCC, we introduce a bi-level warm-up strategy that can stabilize the optimization of the proposed loss function. Our experiments demonstrate that the proposed method significantly outperforms the baselines not only for fine-grained tasks, but also for general vision classification benchmarks.
However, the proposed method can be further improved by optimizing the the strategy used to avoid learning instability. To be precise, the DCC involves contrastive signals from all classes included in a training dataset. Therefore, for a large number of classes, the bi-level warm-up strategy might not be sufficient to control the enormous contrastive signals. Methods such as class subgrouping could remedy this, which will be our focus in future studies. In addition, the proposed method provides potential for further development of dataset condensation by enabling a combination with mixed-class data augmentation methods such as Mixup or CutMix.

%% file: appendix.tex
\section{A Motivating Example}
\setcounter{equation}{5}
\paragraph{An issue with class-wise gradient matching.}
The class-wise gradient matching strategy is employed by previous DC approaches \cite{dc, dsa}. The optimal solution $\tilde{\mathcal{S}}$ of the class-wise gradient matching strategy for Eq. \eqref{eq:ex_matching_loss} can be found as follows:
\begin{equation} \label{eq:class_wise_appendix}
\begin{gathered}
    \tilde{\mathcal{S}}=\argmin_\mathcal{S}\mathcal{L}(\mathcal{X}^+,\mathcal{S}^+)+\mathcal{L}(\mathcal{X}^-,\mathcal{S}^-)
    =
    \argmin_\mathcal{S}\left\| \frac{1}{|\mathcal{X}^+|}\sum_{(x,y)\in\mathcal{X}^+}g_w(x,y)-\frac{1}{|\mathcal{S}^+|}\sum_{(s,t)\in\mathcal{S}^+}g_w(s,t) \right\|\\
    +\left\| \frac{1}{|\mathcal{X}^-|}\sum_{(x,y)\in\mathcal{X}^-}g_w(x,y)-\frac{1}{|\mathcal{S}^-|}\sum_{(s,t)\in\mathcal{S}^-}g_w(s,t) \right\|
    +\frac{\lambda}{|\mathcal{S}^+|}\sum_{s\in\mathcal{S}^+}\|s\|
    +\frac{\lambda}{|\mathcal{S}^-|}\sum_{s\in\mathcal{S}^-}\|s\|\\
     =
    \argmin_\mathcal{S}\left\| \sum_{(x,y)\in\mathcal{X}^+}\frac{-x}{|\mathcal{X}^+|}+\sum_{(s,t)\in\mathcal{S}^+}\frac{s}{|\mathcal{S}^+|} \right\|
    +\left\| \sum_{(x,y)\in\mathcal{X}^-}\frac{x}{|\mathcal{X}^-|}-\sum_{(s,t)\in\mathcal{S}^-}\frac{s}{|\mathcal{S}^-|} \right\|
    +\lambda_\mathcal{S}\\
    =
    \argmin_\mathcal{S}\left\| \mu^+-s_1 \right\|+\left\| \mu^--s_2 \right\|+\lambda_\mathcal{S}
    =
    \left\{\left(\frac{\epsilon\mu^+}{\|\mu^+\|},+1\right), \left(\frac{\epsilon\mu^-}{\|\mu^-\|},-1\right)\right\},
\end{gathered}
\end{equation}
where $\mu^+=\frac{1}{|\mathcal{X}^+|}\sum_{x\in\mathcal{X}^+}x$, $\mu^-=\frac{1}{|\mathcal{X}^-|}\sum_{x\in\mathcal{X}^-}x$, and $\lambda_\mathcal{S}=\lambda\sum_{s\in\mathcal{S}}\|s\|$.
Eq \eqref{eq:class_wise_appendix} demonstrates that the class-wise gradient matching method optimizes $\mathcal{S}$, for each class, to have the same direction as the average of training samples that generate gradients.
Then, $R(\tilde{\mathcal{S}})$ is as follows:
\begin{equation}\label{eq:R_class_wise_appendix}
R\left( \tilde{\mathcal{S}} \right)
=
\frac{1}{2}\left(
\frac{\abs{\frac{\phi_1^\top\mu^+}{\| \mu^+ \|}}}{\abs{\frac{\phi_1^\top\mu^+}{\| \mu^+ \|}}+\abs{\frac{\phi_2^\top\mu^+}{\| \mu^+ \|}}}
+\frac{\abs{\frac{\phi_1^\top\mu^-}{\| \mu^- \|}}}{\abs{\frac{\phi_1^\top\mu^-}{\| \mu^- \|}}+\abs{\frac{\phi_2^\top\mu^-}{\| \mu^- \|}}}
\right)
=
\frac{\abs{\frac{\phi_1^\top\mu^+}{\| \mu^+ \|}}}{\abs{\frac{\phi_1^\top\mu^+}{\| \mu^+ \|}}+\beta}
\leq\frac{\alpha}{\alpha+\beta}.
\end{equation}
Here, without loss of generality, we assume the cardinality of $\mathcal{X}$ is sufficiently large such that $\abs{\frac{\phi_1^\top\mu^+}{\| \mu^+ \|}}=\abs{\frac{\phi_1^\top\mu^-}{\| \mu^- \|}}$. The equality holds when $\beta=0$, and the inequality is due to $\abs{\frac{\phi_1^\top\mu^+}{\| \mu^+ \|}}<\alpha$.
Eq. \eqref{eq:R_class_wise_appendix} shows that when $\alpha \ll \beta$ (\ie, when class-common features are dominant features and class-discriminative features are minority), $R(\tilde S) \rightarrow 0$, \ie, the class-wise gradient matching method can result in synthetic datasets that are ineffectual for the classification task.
For example, as shown in Table \ref{tab:finegrained}, the class-wise gradient matching method can fail on fine-grained classification tasks that include shared appearance between classes, and can be discriminative by only fine-grained appearances.

\paragraph{Leveraging contrastive signals.}
The class-wise gradient matching method has a limitation when the class-common features are dominant. We need a different approach to capture only class-discriminative features for better target task performances. The following simple modification of Eq. \eqref{eq:class_wise_appendix} can mitigate the issue:
\begin{equation} \label{eq:class_comprehensive_appendix}
    \begin{gathered}
    \hat{\mathcal{S}}
    =
    \argmin_\mathcal{S}\mathcal{L}(\mathcal{X}^+\cup\mathcal{X}^-,\mathcal{S}^+\cup\mathcal{S}^-)\\
    =
    \argmin_\mathcal{S}\left\| \frac{1}{|\mathcal{X}^+|+|\mathcal{X}^-|}\sum_{(x,y)\in\mathcal{X}^+\cup\mathcal{X}^-}g_w(x,y)-\frac{1}{|\mathcal{S}^+|+|\mathcal{S}^-|}\sum_{(s,t)\in\mathcal{S}^+\cup\mathcal{S}^-}g_w(s,t) \right\|+\lambda_{\mathcal{S}}\\
    =
    \argmin_\mathcal{S}\left\| \left(\sum_{(x,y)\in\mathcal{X}^+}\frac{-x}{2\bar{N}}+\sum_{(x,y)\in\mathcal{X}^-}\frac{x}{2\bar{N}}\right)-\left(\sum_{(s,t)\in\mathcal{S}^+}\frac{-s}{2}+\sum_{(s,t)\in\mathcal{S}^-}\frac{s}{2}\right) \right\|+\lambda_{\mathcal{S}}\\
    =
    \argmin_\mathcal{S}\left\| (\mu^+-\mu^-)-(s_1-s_2) \right\|+\lambda_\mathcal{S}
    =
    \left\{\left(\epsilon\phi_1,+1\right), \left(-\epsilon\phi_1,-1\right)\right\}.
\end{gathered}
\end{equation}
In Eq. \eqref{eq:setup}, we can see that $\mathcal{X}$ is balanced ($y\stackrel{u.a.r}{\sim} \{-1, +1\}$). Hence, we can set $|\mathcal{X}^+|=|\mathcal{X}^-|=\hat{N}$, without loss of generality. Eq. \eqref{eq:class_comprehensive_appendix} considers loss gradients for all classes collectively, while Eq. \eqref{eq:class_wise_appendix} considers loss gradients for each class separately. Moreover, Eq. \eqref{eq:class_comprehensive_appendix} reveals that the sum of loss gradients between classes is important because it contains contrastive signals between classes ($(\mu^+-\mu^-)$ and $(s_1-s_2)$).
Here, $R(\hat{\mathcal{S}})$ is as follows:
\begin{equation}\label{eq:R_class_collective_appendix}
R\left( \hat{\mathcal{S}} \right)
=
\frac{1}{2}\left( \frac{2\epsilon|\phi_1^\top\phi_1|}{\epsilon|\phi_1^\top\phi_1|+\epsilon|\phi_1^\top\phi_2|}\right)
=
1,
\end{equation}
in other words, $\hat{\mathcal{S}}$ contains only class-discriminative features, so that it is independent of the proportion of class-common features in the original training dataset $\mathcal{X}$.

\section{Application: Continual Learning}
\input{Figure/figure_continual_cifar}
Figure \ref{fig:continual_cifar} shows the effectiveness of DSAC, DSA and RB on the continual learning task which is composed of CIFAR10, SVHN and TrafficSigns \cite{stallkamp2011german}. DSAC and DSA utilize the condensed datasets as rehearsal examples as in Figure \ref{fig:continual}. We note that our DSAC again dominates other baselines for T2 and T3 tasks.

\section{Datasets}
\input{Table/table_subsampled_classes}
SVHN~\cite{svhn_dataset} consists of 73,257 training images and 26,032 test images in 10 classes.
CIFAR-10~\cite{cifar_dataset} consists of 50,000 training images and 10,000 test images in 10 classes.
CIFAR-100~\cite{cifar_dataset} consists of 50,000 training images and 10,000 test images in 100 classes.
SVHN, CIFAR-10, and CIFAR-100 images have sizes of $32\times32$ pixels.
ImageNet~\cite{imagenet_dataset} consists of 1,281,167 training images and 100,000 test images in 1,000 classes.
\citet{downsampled_imagenet} provided downsampled variants of the ImageNet dataset. The ImageNet32x32 dataset~\cite{downsampled_imagenet} have the same number of classes and images as ImageNet, but the images are downsampled to sizes of $32\times32$ pixel.
We constructed the fine-grained image classification datasets by subsampling from the ImageNet32x32 dataset using the WordNet \cite{wordnet} hierarchy. The subsampled ImageNet classes are summarized in Tab.~\ref{tab:table_subsampled_classes}.

%% file: Figure/figure_continual_cifar.tex
\begin{figure}[t!]
{
\begin{center}
\centerline{\includegraphics[width=0.5\columnwidth]{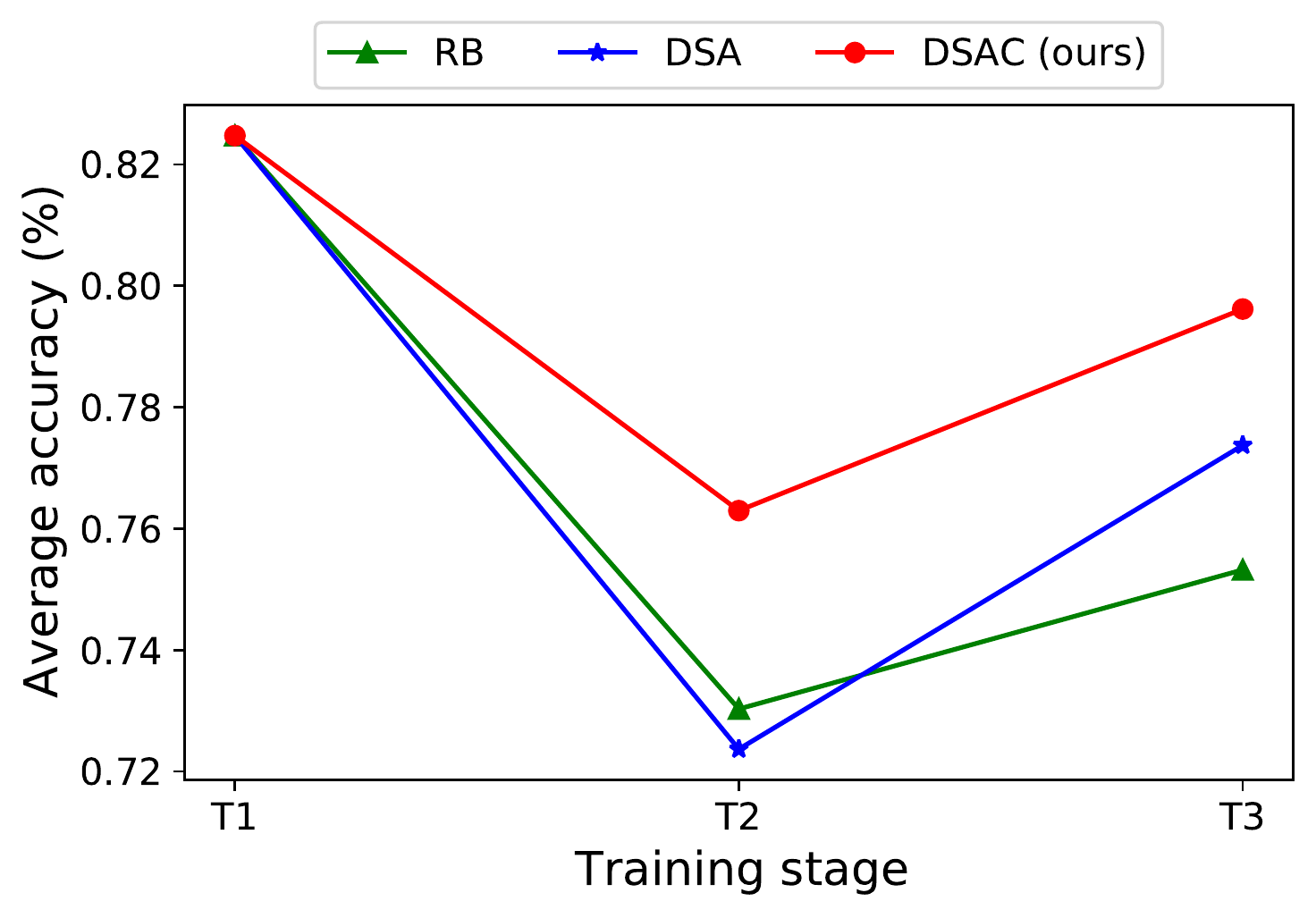}}
\caption{Performance improvements (average accuracy \%) on the continual learning task for a sequence of benchmark datasets \{CIFAR10-SVHN-TrafficSigns\} following the application of the proposed method.}
\label{fig:continual_cifar}
\end{center}
}
\end{figure}

%% file: Table/table_subsampled_classes.tex
\begin{table*}[]
\centering
\caption{The subsampled ImageNet classes for each fine-grained image classification dataset.}
\label{tab:table_subsampled_classes}
\small
\begin{tabular}{@{}ll@{}}
\toprule
Fine-grained dataset & ImageNet classes \\ \midrule
Automobile & beach wagon, convertible, sports   car, ambulance, jeep, limousine, racer, cab, Model T \\
&\\
Terrier &  Lakeland terrier, Scotch terrier, cairn, Airedale, Tibetan terrier, Yorkshire terrier, Norfolk terrier,\\
&Staffordshire bullterrier, Sealyham terrier, standard schnauzer, Norwich terrier, Bedlington terrier, Lhasa,\\
&Irish terrier, silky terrier,  Dandie Dinmont, Boston bull, Border terrier,soft-coated wheaten terrier,\\
&Australian terrier, American Staffordshire terrier, West Highland white terrier, giant schnauzer,\\
&miniature schnauzer, Kerry blue terrier, wire-haired fox terrier \\
&\\
Fish & tench, stingray, tiger shark, barracouta, coho, gar, electric ray, great white shark, sturgeon\\
&puffer, anemone fish, goldfish, eel, rock beauty, lionfish, hammerhead \\
&\\
Lizard & agama, banded gecko, Komodo dragon, frilled lizard, African chameleon, American chameleon,\\
&green lizard, whiptail, common iguana, alligator lizard, Gila monster \\
&\\
Truck & pickup, police van, trailer truck,   minivan, moving van, tow truck, fire engine, garbage truck, tractor \\
&\\
Insect & cricket, ant, leafhopper, walking stick, grasshopper, dung beetle, tiger beetle, lacewing,\\
&rhinoceros beetle, ringlet, long-horned beetle, ladybug, ground beetle, cicada, cabbage butterfly,\\
&leaf beetle, lycaenid, bee, monarch, damselfly, admiral, sulphur butterfly, dragonfly, fly, weevil,\\
&cockroach, mantis \\\bottomrule
\end{tabular}
\end{table*}

%% file: main.bbl
\begin{thebibliography}{36}
\providecommand{\natexlab}[1]{#1}
\providecommand{\url}[1]{\texttt{#1}}
\expandafter\ifx\csname urlstyle\endcsname\relax
  \providecommand{\doi}[1]{doi: #1}\else
  \providecommand{\doi}{doi: \begingroup \urlstyle{rm}\Url}\fi

\bibitem[Brown et~al.(2020)Brown, Mann, Ryder, Subbiah, Kaplan, Dhariwal,
  Neelakantan, Shyam, Sastry, Askell, et~al.]{gpt3}
Brown, T.~B., Mann, B., Ryder, N., Subbiah, M., Kaplan, J., Dhariwal, P.,
  Neelakantan, A., Shyam, P., Sastry, G., Askell, A., et~al.
\newblock Language models are few-shot learners.
\newblock \emph{arXiv preprint arXiv:2005.14165}, 2020.

\bibitem[Chaudhry et~al.(2019)Chaudhry, Rohrbach, Elhoseiny, Ajanthan, Dokania,
  Torr, and Ranzato]{chaudhry2019tiny}
Chaudhry, A., Rohrbach, M., Elhoseiny, M., Ajanthan, T., Dokania, P.~K., Torr,
  P.~H., and Ranzato, M.
\newblock On tiny episodic memories in continual learning.
\newblock \emph{arXiv preprint arXiv:1902.10486}, 2019.

\bibitem[Chrabaszcz et~al.(2017)Chrabaszcz, Loshchilov, and
  Hutter]{downsampled_imagenet}
Chrabaszcz, P., Loshchilov, I., and Hutter, F.
\newblock A downsampled variant of imagenet as an alternative to the cifar
  datasets.
\newblock \emph{arXiv preprint arXiv:1707.08819}, 2017.

\bibitem[Deng et~al.(2009)Deng, Dong, Socher, Li, Li, and
  Fei-Fei]{imagenet_dataset}
Deng, J., Dong, W., Socher, R., Li, L.-J., Li, K., and Fei-Fei, L.
\newblock {ImageNet: A Large-Scale Hierarchical Image Database}.
\newblock In \emph{CVPR09}, 2009.

\bibitem[Fort et~al.(2020)Fort, Dziugaite, Paul, Kharaghani, Roy, and
  Ganguli]{kernel_velocity}
Fort, S., Dziugaite, G.~K., Paul, M., Kharaghani, S., Roy, D.~M., and Ganguli,
  S.
\newblock Deep learning versus kernel learning: an empirical study of loss
  landscape geometry and the time evolution of the neural tangent kernel.
\newblock \emph{Advances in Neural Information Processing Systems}, 33, 2020.

\bibitem[Gidaris \& Komodakis(2018)Gidaris and Komodakis]{convnet}
Gidaris, S. and Komodakis, N.
\newblock Dynamic few-shot visual learning without forgetting.
\newblock In \emph{Proceedings of the IEEE Conference on Computer Vision and
  Pattern Recognition}, pp.\  4367--4375, 2018.

\bibitem[Goodfellow et~al.(2013)Goodfellow, Mirza, Xiao, Courville, and
  Bengio]{catastrophic_forgetting}
Goodfellow, I.~J., Mirza, M., Xiao, D., Courville, A., and Bengio, Y.
\newblock An empirical investigation of catastrophic forgetting in
  gradient-based neural networks.
\newblock \emph{arXiv preprint arXiv:1312.6211}, 2013.

\bibitem[He et~al.(2016)He, Zhang, Ren, and Sun]{resnet}
He, K., Zhang, X., Ren, S., and Sun, J.
\newblock Deep residual learning for image recognition.
\newblock In \emph{Proceedings of the IEEE conference on computer vision and
  pattern recognition}, pp.\  770--778, 2016.

\bibitem[Hern{\'a}ndez-Garc{\'\i}a \& K{\"o}nig(2018)Hern{\'a}ndez-Garc{\'\i}a
  and K{\"o}nig]{da_reg}
Hern{\'a}ndez-Garc{\'\i}a, A. and K{\"o}nig, P.
\newblock Data augmentation instead of explicit regularization.
\newblock \emph{arXiv preprint arXiv:1806.03852}, 2018.

\bibitem[Jastrzebski et~al.(2020)Jastrzebski, Szymczak, Fort, Arpit, Tabor,
  Cho, and Geras]{break-even_point}
Jastrzebski, S., Szymczak, M., Fort, S., Arpit, D., Tabor, J., Cho, K., and
  Geras, K.
\newblock The break-even point on optimization trajectories of deep neural
  networks.
\newblock \emph{arXiv preprint arXiv:2002.09572}, 2020.

\bibitem[Jia et~al.(2021)Jia, Yang, Xia, Chen, Parekh, Pham, Le, Sung, Li, and
  Duerig]{align_google}
Jia, C., Yang, Y., Xia, Y., Chen, Y.-T., Parekh, Z., Pham, H., Le, Q.~V., Sung,
  Y., Li, Z., and Duerig, T.
\newblock Scaling up visual and vision-language representation learning with
  noisy text supervision.
\newblock \emph{arXiv preprint arXiv:2102.05918}, 2021.

\bibitem[Jiang et~al.(2021)Jiang, Zhang, Talwar, and Mozer]{cscores}
Jiang, Z., Zhang, C., Talwar, K., and Mozer, M.~C.
\newblock Characterizing structural regularities of labeled data in
  overparameterized models.
\newblock In Meila, M. and Zhang, T. (eds.), \emph{Proceedings of the 38th
  International Conference on Machine Learning}, volume 139 of
  \emph{Proceedings of Machine Learning Research}, pp.\  5034--5044. PMLR,
  18--24 Jul 2021.
\newblock URL \url{https://proceedings.mlr.press/v139/jiang21k.html}.

\bibitem[Kim et~al.(2018)Kim, Kim, Seo, Kim, Park, Park, Jo, Kim, Yang, Kim,
  et~al.]{nsml2}
Kim, H., Kim, M., Seo, D., Kim, J., Park, H., Park, S., Jo, H., Kim, K., Yang,
  Y., Kim, Y., et~al.
\newblock Nsml: Meet the mlaas platform with a real-world case study.
\newblock \emph{arXiv preprint arXiv:1810.09957}, 2018.

\bibitem[Krizhevsky et~al.(2009)Krizhevsky, Hinton, et~al.]{cifar_dataset}
Krizhevsky, A., Hinton, G., et~al.
\newblock Learning multiple layers of features from tiny images.
\newblock Technical report, Citeseer, 2009.

\bibitem[Krizhevsky et~al.(2012)Krizhevsky, Sutskever, and Hinton]{alexnet}
Krizhevsky, A., Sutskever, I., and Hinton, G.~E.
\newblock Imagenet classification with deep convolutional neural networks.
\newblock \emph{Advances in neural information processing systems},
  25:\penalty0 1097--1105, 2012.

\bibitem[LeCun(1998)]{mnist_dataset}
LeCun, Y.
\newblock The mnist database of handwritten digits.
\newblock \emph{http://yann. lecun. com/exdb/mnist/}, 1998.

\bibitem[LeCun et~al.(1998)LeCun, Bottou, Bengio, and Haffner]{lenet}
LeCun, Y., Bottou, L., Bengio, Y., and Haffner, P.
\newblock Gradient-based learning applied to document recognition.
\newblock \emph{Proceedings of the IEEE}, 86\penalty0 (11):\penalty0
  2278--2324, 1998.

\bibitem[Lee et~al.(2021)Lee, Park, Lee, Yi, Lee, and Yoon]{oat}
Lee, S., Park, C., Lee, H., Yi, J., Lee, J., and Yoon, S.
\newblock Removing undesirable feature contributions using out-of-distribution
  data.
\newblock In \emph{International Conference on Learning Representations}, 2021.
\newblock URL \url{https://openreview.net/forum?id=eIHYL6fpbkA}.

\bibitem[Liu et~al.(2020)Liu, Jiang, Bai, Chen, and Wang]{gsnr}
Liu, J., Jiang, G., Bai, Y., Chen, T., and Wang, H.
\newblock Understanding why neural networks generalize well through gsnr of
  parameters.
\newblock \emph{arXiv preprint arXiv:2001.07384}, 2020.

\bibitem[Mahajan et~al.(2018)Mahajan, Girshick, Ramanathan, He, Paluri, Li,
  Bharambe, and Van Der~Maaten]{instagramnet}
Mahajan, D., Girshick, R., Ramanathan, V., He, K., Paluri, M., Li, Y.,
  Bharambe, A., and Van Der~Maaten, L.
\newblock Exploring the limits of weakly supervised pretraining.
\newblock In \emph{Proceedings of the European conference on computer vision
  (ECCV)}, pp.\  181--196, 2018.

\bibitem[Miller(1998)]{wordnet}
Miller, G.~A.
\newblock \emph{WordNet: An electronic lexical database}.
\newblock MIT press, 1998.

\bibitem[Mirzasoleiman et~al.(2020)Mirzasoleiman, Bilmes, and
  Leskovec]{coresets_for_efficient}
Mirzasoleiman, B., Bilmes, J., and Leskovec, J.
\newblock Coresets for data-efficient training of machine learning models.
\newblock In III, H.~D. and Singh, A. (eds.), \emph{Proceedings of the 37th
  International Conference on Machine Learning}, volume 119 of
  \emph{Proceedings of Machine Learning Research}, pp.\  6950--6960. PMLR,
  13--18 Jul 2020.
\newblock URL \url{https://proceedings.mlr.press/v119/mirzasoleiman20a.html}.

\bibitem[Netzer et~al.(2011)Netzer, Wang, Coates, Bissacco, Wu, and
  Ng]{svhn_dataset}
Netzer, Y., Wang, T., Coates, A., Bissacco, A., Wu, B., and Ng, A.~Y.
\newblock Reading digits in natural images with unsupervised feature learning.
\newblock 2011.

\bibitem[Nguyen et~al.(2020)Nguyen, Chen, and Lee]{kip}
Nguyen, T., Chen, Z., and Lee, J.
\newblock Dataset meta-learning from kernel ridge-regression.
\newblock In \emph{International Conference on Learning Representations}, 2020.

\bibitem[Nguyen et~al.(2021)Nguyen, Novak, Xiao, and Lee]{dd_infinite}
Nguyen, T., Novak, R., Xiao, L., and Lee, J.
\newblock Dataset distillation with infinitely wide convolutional networks.
\newblock In Beygelzimer, A., Dauphin, Y., Liang, P., and Vaughan, J.~W.
  (eds.), \emph{Advances in Neural Information Processing Systems}, 2021.
\newblock URL \url{https://openreview.net/forum?id=hXWPpJedrVP}.

\bibitem[Paul et~al.(2021)Paul, Ganguli, and Dziugaite]{diet}
Paul, M., Ganguli, S., and Dziugaite, G.~K.
\newblock Deep learning on a data diet: Finding important examples early in
  training.
\newblock \emph{Advances in Neural Information Processing Systems}, 34, 2021.

\bibitem[Rebuffi et~al.(2017)Rebuffi, Kolesnikov, Sperl, and
  Lampert]{rebuffi2017icarl}
Rebuffi, S.-A., Kolesnikov, A., Sperl, G., and Lampert, C.~H.
\newblock icarl: Incremental classifier and representation learning.
\newblock In \emph{Proceedings of the IEEE conference on Computer Vision and
  Pattern Recognition}, pp.\  2001--2010, 2017.

\bibitem[Simonyan \& Zisserman(2014)Simonyan and Zisserman]{vgg}
Simonyan, K. and Zisserman, A.
\newblock Very deep convolutional networks for large-scale image recognition.
\newblock \emph{arXiv preprint arXiv:1409.1556}, 2014.

\bibitem[Stallkamp et~al.(2011)Stallkamp, Schlipsing, Salmen, and
  Igel]{stallkamp2011german}
Stallkamp, J., Schlipsing, M., Salmen, J., and Igel, C.
\newblock The german traffic sign recognition benchmark: a multi-class
  classification competition.
\newblock In \emph{The 2011 international joint conference on neural networks},
  pp.\  1453--1460. IEEE, 2011.

\bibitem[Sung et~al.(2017)Sung, Kim, Jo, Yang, Kim, Lausen, Kim, Lee, Kwak, Ha,
  et~al.]{nsml1}
Sung, N., Kim, M., Jo, H., Yang, Y., Kim, J., Lausen, L., Kim, Y., Lee, G.,
  Kwak, D., Ha, J.-W., et~al.
\newblock Nsml: A machine learning platform that enables you to focus on your
  models.
\newblock \emph{arXiv preprint arXiv:1712.05902}, 2017.

\bibitem[Wang \& Isola(2020)Wang and Isola]{wang2020uniformity}
Wang, T. and Isola, P.
\newblock Understanding contrastive representation learning through alignment
  and uniformity on the hypersphere.
\newblock In \emph{International Conference on Machine Learning}, pp.\
  9929--9939. PMLR, 2020.

\bibitem[Wang et~al.(2018)Wang, Zhu, Torralba, and Efros]{dd}
Wang, T., Zhu, J.-Y., Torralba, A., and Efros, A.~A.
\newblock Dataset distillation.
\newblock \emph{arXiv preprint arXiv:1811.10959}, 2018.

\bibitem[Yun et~al.(2019)Yun, Han, Oh, Chun, Choe, and Yoo]{cutmix}
Yun, S., Han, D., Oh, S.~J., Chun, S., Choe, J., and Yoo, Y.
\newblock Cutmix: Regularization strategy to train strong classifiers with
  localizable features.
\newblock In \emph{Proceedings of the IEEE International Conference on Computer
  Vision}, pp.\  6023--6032, 2019.

\bibitem[Zhang et~al.(2017)Zhang, Cisse, Dauphin, and Lopez-Paz]{mixup}
Zhang, H., Cisse, M., Dauphin, Y.~N., and Lopez-Paz, D.
\newblock mixup: Beyond empirical risk minimization.
\newblock \emph{arXiv preprint arXiv:1710.09412}, 2017.

\bibitem[Zhao \& Bilen(2021)Zhao and Bilen]{dsa}
Zhao, B. and Bilen, H.
\newblock Dataset condensation with differentiable siamese augmentation.
\newblock In \emph{International Conference on Machine Learning}, 2021.

\bibitem[Zhao et~al.(2021)Zhao, Mopuri, and Bilen]{dc}
Zhao, B., Mopuri, K.~R., and Bilen, H.
\newblock Dataset condensation with gradient matching.
\newblock In \emph{International Conference on Learning Representations}, 2021.
\newblock URL \url{https://openreview.net/forum?id=mSAKhLYLSsl}.

\end{thebibliography}
